\PassOptionsToPackage{table}{xcolor}
\documentclass[sigconf]{acmart}

\AtBeginDocument{%
  }

\usepackage[table]{xcolor}
\usepackage{multirow, soul}
\usepackage{amsmath}

\DeclareMathOperator*{\argmin}{arg\,min}

\copyrightyear{2025}
\acmYear{2025}
\setcopyright{cc}
\setcctype{by-nc}
\acmConference[MM '25]{Proceedings of the 33rd ACM International Conference on Multimedia}{October 27--31, 2025}{Dublin, Ireland}
\acmBooktitle{Proceedings of the 33rd ACM International Conference on Multimedia (MM '25), October 27--31, 2025, Dublin, Ireland}\acmDOI{10.1145/3746027.3755763}
\acmISBN{979-8-4007-2035-2/2025/10}

\newcommand{\SystemName}{PromptFlare}

\definecolor{backcolour}{RGB}{230, 230, 230}

\begin{document}

\title{PromptFlare: Prompt-Generalized Defense via Cross-Attention Decoy in Diffusion-Based Inpainting}

\author{Hohyun Na}
\affiliation{%
    \institution{Sungkyunkwan University}
    \department{Department of Intelligent Software}
    \city{Suwon}
    \country{Republic of Korea}
}
\email{skghgus9@g.skku.edu}

\author{Seunghoo Hong}
\affiliation{%
    \institution{Sungkyunkwan University}
    \department{Department of Artificial Intelligence}
    \city{Suwon}
    \country{Republic of Korea}
}
\email{hoo0681@g.skku.edu}

\author{Simon S. Woo}
\authornote{Corresponding author.}
\affiliation{%
    \institution{Sungkyunkwan University}
    \department{Department of Computer Science and Engineering}
    \city{Suwon}
    \country{Republic of Korea}
}
\email{swoo@g.skku.edu}

\begin{abstract}
The success of diffusion models has enabled effortless, high-quality image modifications that precisely align with users' intentions, thereby raising concerns about their potential misuse by malicious actors. Previous studies have attempted to mitigate such misuse through adversarial attacks. However, these approaches heavily rely on image-level inconsistencies, which pose fundamental limitations in addressing the influence of textual prompts. In this paper, we propose \SystemName, a novel adversarial protection method designed to protect images from malicious modifications facilitated by diffusion-based inpainting models. Our approach leverages the cross-attention mechanism to exploit the intrinsic properties of prompt embeddings. Specifically, we identify and target shared token of prompts that is invariant and semantically uninformative, injecting adversarial noise to suppress the sampling process. The injected noise acts as a cross-attention decoy, diverting the model’s focus away from meaningful prompt-image alignments and thereby neutralizing the effect of prompt. Extensive experiments on the EditBench dataset demonstrate that our method achieves state-of-the-art performance across various metrics while significantly reducing computational overhead and GPU memory usage. These findings highlight \SystemName~ as a robust and efficient protection against unauthorized image manipulations. The code is available at https://github.com/NAHOHYUN-SKKU/PromptFlare.
\end{abstract}



\maketitle

\section{Introduction}
\label{sec: Introduction}

\begin{figure}[t]
  \centering
  \includegraphics[width=0.95\linewidth]{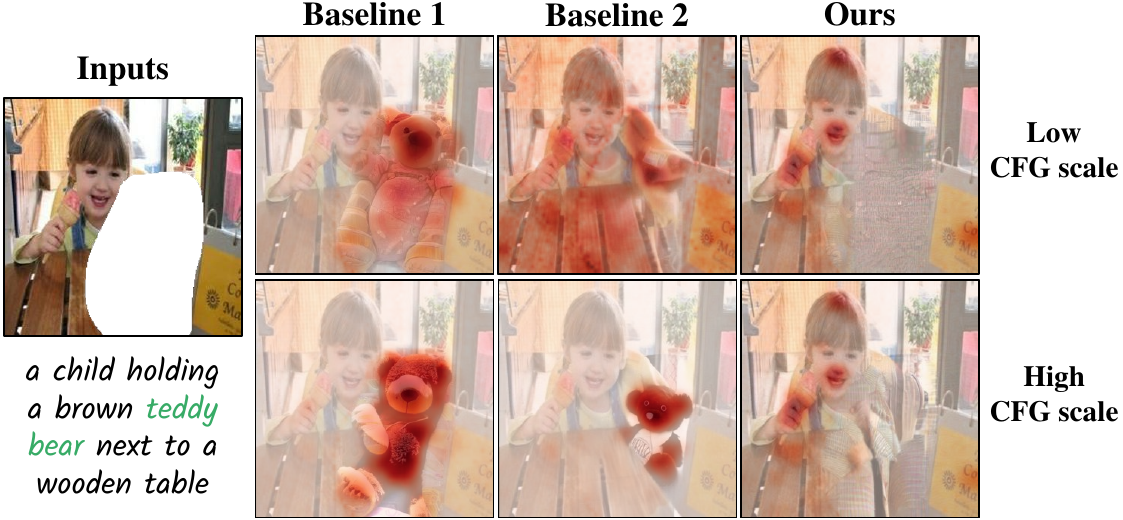}
  \caption{Inpainting results with a masked image and prompt. The green word denotes the core object intended for generation, with its attention map overlaid on the output image. Our method consistently suppresses generation across all CFG scales by removing the prompt’s influence.}
  \Description{The input image and the Inpainting results of baseline and our method are displayed.}
  \label{Fig: introduction}
\end{figure}

Recently, diffusion-based generative models~\cite{ho2020denoising,song2020denoising,saharia2022photorealistic}, such as Stable Diffusion~\cite{rombach2022high} and DALL·E 2~\cite{ramesh2022hierarchical} have substantially enhanced AI-driven image synthesis and editing techniques~\cite{mokady2023null, wallace2023edict, Xie_2023_CVPR,meng2021sdedit,hertz2022prompt}. These models enable users to generate high-quality, realistic images from text descriptions and modify existing images with remarkable precision. In particular, diffusion-based inpainting models have gained popularity for their ability to seamlessly reconstruct masked regions while preserving visual coherence. This has been widely applied to tasks such as object replacement and image editing~\cite{wang2023imagen,yenphraphai2024image,brooks2023instructpix2pix}. However, as such models have become more accessible, concerns have emerged for their potential misuse~\cite{li2022seeing,tariq2023deepfake}. Malicious users can exploit inpainting models to alter images deceptively, such as modifying personal photos without consent, removing watermarks, or generating misleading visual content. While traditional image editing techniques require manual effort and expertise, diffusion-based inpainting lowers the barrier to image manipulation by simple prompting, and masking. 

Previous studies have attempted to mitigate these threats by protecting images through adversarial attacks that inject noise to disrupt the internal mechanisms of diffusion models. For example, PhotoGuard~\cite{salman2023raising}, DDD~\cite{son2024disrupting}, AdvPaint~\cite{jeon2025advpaint} focus on degrading the generation process, and DiffusionGuard~\cite{choi2024diffusionguard} extends these methods to more complex editing scenarios. A common limitation of these approaches is that they structure the loss function solely based on image-level inconsistencies, overlooking the critical role of prompts in guiding the inpainting process. As shown in Fig. \ref{Fig: introduction}, a malicious user can easily circumvent these defenses by simply increasing the guidance scale, a tunable parameter, to generate images that remain faithful to the prompts.

\SystemName~ addresses this limitation by explicitly integrating the semantic influence of textual prompts into the adversarial objective. We generate the adversarial noise ensuring that the image attends only from the meaningless common part of the prompts while ignoring the rest. Specifically, we utilize an attention mask to isolate shared tokens that are common across all prompts when images and prompts interact via cross-attention. We target the cross-attention output with mask and introduce a loss term that minimizes the discrepancy compared to the output obtained without the mask. In this framework, adversarial noise acts as a decoy within the cross-attention mechanism, impairing model's ability to align the malicious prompts with images. As a result, this strategy effectively disrupts the generation process, preventing the model from producing images faithful to malicious prompts and neutralizing exploitability regardless of the CFG scale or other parameters. The effectiveness of this approach is further supported by Fig. \ref{Fig: introduction}, which presents attention maps over core object tokens and visually confirms that our method successfully inhibits impact of prompt.

\SystemName~ introduces a concise yet effective objective function by precisely aligning the goal with the objective function, resulting in minimal computational overhead. By integrating directly into the diffusion-based inpainting pipeline and focusing on the critical interactions between prompts and images, our approach enables robust, prompt-generalized adversarial noise generation without requiring additional resources. Through a comprehensive evaluation, our method achieved state-of-the-art (SoTA) performance across most metrics. 
Furthermore, it operates with substantially lower GPU memory requirements and reduces runtime to roughly 1/3 compared to prior methods. This efficiency, combined with its straightforward implementation, ensures high scalability and accessibility on consumer-grade hardware, thereby extending protection opportunities to a broader range of users.

Our key contributions are summarized as follows:

\begin{itemize}
    \item We introduce \SystemName, a novel adversarial defense framework that suppresses malicious inpainting by targeting cross-attention on prompt embeddings.
    \item Our approach is prompt-generalized, focusing on the shared token of prompts to neutralize the influence of malicious prompts.
    \item Extensive evaluations conducted on a comprehensive dataset demonstrate that our method achieves superior performance across metrics, significantly reduces computational overhead, and enhances scalability and practical applicability.
\end{itemize}
\section{Related Work}
\label{Sec: RelatedWork}
\paragraph{Stable Diffusion models} Stable Diffusion (SD)~\cite{rombach2022high} is a representative Latent Diffusion Model (LDM) designed to efficiently produce high-quality images by operating within a latent space rather than the pixel space. As one of the most widely adopted open source LDMs, Stable Diffusion has been extensively used for both research and practical applications. During the image generation process, it leverages a pre-trained noise prediction model to perform the reverse diffusion process, progressively reconstructing meaningful structures from random noise~\cite{ho2020denoising, sohl2015deep, dhariwal2021diffusion}. 
Stable Diffusion supports Text-to-Image generation by integrating CLIP’s text encoder~\cite{radford2021learning}, translating textual prompts into latent representations to synthesize corresponding visual outputs~\cite{sauer2024fast, saharia2022photorealistic, betker2023improving}.

\paragraph{Inpainting-based generation suppression}
Previous studies have employed inpainting models to inject adversarial noise, aiming to prevent malevolent users from successfully performing their intended inpainting modifications. PhotoGuard~\cite{salman2023raising} is the first work to introduce the generation suppression in diffusion models, employing adversarial attacks to steer the reverse diffusion process away from intended image generation.
DDD ~\cite{son2024disrupting} proposed a framework that integrates PEZ~\cite{wen2023hard} to extract textual prompts that effectively represent the semantic information of an image~\cite{gal2022image}.
DiffusionGuard~\cite{choi2024diffusionguard} 
repeatedly applied adversarial attacks on randomly and iteratively shrunk masks, forcing the generated image to move away from itself, thereby mitigating the overfitting issue.
AdvPaint~\cite{jeon2025advpaint} exploits self-attention and cross-attention about the only image and pushes it away by considering the attention layer as an intermediate feature. However, previous approaches focus solely on image-level inconsistencies, making them highly vulnerable to suppression failure when the influence of the prompt is amplified during the inpainting process. In this paper, we analyze how prompts are incorporated into the inpainting process and aim to directly suppress their influence.

\paragraph{Editing-based generation suppression.} Image editing is another application domain wherein Stable Diffusion is employed to modify input images. Previous studies such as PhotoGuard~\cite{salman2023raising}, AdvDM~\cite{liang2023adversarial}, Mist~\cite{liang2023mist} and SDS~\cite{xuetoward} have been proposed to prevent malicious editing in this task. While editing and inpainting share similarities, a key distinction lies in the absence of an explicit mask in the former. Furthermore, editing typically introduces Gaussian noise to the input image, which serves as a reference during generation. This fundamental difference necessitates distinct strategies for generation suppression across the two tasks. Although PhotoGuard addressed both tasks, it just separately uses two models rather than integrating adversarial noise. 
\section{Background}
\label{Sec:Background}

\subsection{Stable Diffusion Inpainting}
\label{Subsec: Inpainting background}

\paragraph{Analysis on inpainting} The Stable Diffusion inpainting model is a variant of Stable Diffusion that has been fine-tuned specifically for inpainting tasks~\cite{rombach2022high}. It consists of a forward process and a reverse diffusion process. In the forward process, Gaussian noise is gradually added to the image latent vector $z_0$ until it reaches $z_{T}$. In the reverse diffusion process, the noise prediction model is used to iteratively remove the estimated noise from $z_T$, progressively reconstructing $z_0^\prime$. For the forward process, given an image $x \in \mathbb{R}^{3 \times W \times H}$, Variational AutoEncoder (VAE)~\cite{van2017neural} encoder $\mathcal{E}$, and the corresponding image latent vector $z_0=\mathcal{E}(x)\in \mathbb{R}^{4\times w\times h}$, the latent representation $z_t$ at timestamp $t$ is computed as follows:
    
    
\begin{align}
\label{Eq: forward process in inpainting}
q(z_t \mid z_{t-1}) &:= \mathcal{N}(z_t;\sqrt{\alpha_t}z_{t-1}, (1-\alpha_t)\mathbf{I}), \\
q(z_T \mid z_0) &\approx \mathcal{N}(z_T;\mathbf{0},\mathbf{I}),\nonumber
\end{align}


where $T$ is the number of timestep, and $z_T$ typically  becomes pure Gaussian noise.

The reverse diffusion process involves predicting and removing this noise step-by-step using an inpainting-specific noise prediction model $\epsilon_{\theta}$. This model leverages a mask $\mathbf{M}\in \mathbb{R}^{1\times W \times H}$ to distinguish between areas of the image to retain and those requiring inpainting. The mask-applied latent representation $z^\mathbf{M}_0$ is computed as:

\begin{equation}
\label{Eq: image context}
    z^M_0=\mathcal{E}(x\otimes \mathbf{M} ), ~~\text{where} ~~ M_{i,j} = 
    \begin{cases}
        1, & \text{if to be maintain} \\
        0, & \text{if to be inpaint}
    \end{cases}.
\end{equation}
The noise prediction model $\epsilon_{\theta}$ is trained with the following loss function, minimizing the difference between the predicted noise and the actual noise added:
\begin{equation}
\label{Eq: inpainting loss term}
    L_{\text{inpainting}}=\mathbb{E}_{z_t,t \sim [1,T],\epsilon \sim \mathcal{N}(0,1),}||\epsilon - \epsilon_\theta ( \tilde z_t,t, \mathbf{e}) ||_2,
\end{equation}
where $\tilde z_t$ is $\text{concat}(z_t. \mathbf{M}^\prime , z^\mathbf{M}_0)$, $\mathbf{M}^\prime \in \mathbb{R}^{1\times w \times h}$ is the downsampled $\mathbf{M}$ aligned with latent dimensions and $\mathbf{e}$ is prompt embedding. As image-level context, the noise prediction model utilizes the downsampled mask $\mathbf{M}^\prime$ and the clean masked latent $z^{\mathbf{M}}_0$. These are concatenated with the currently generated noisy image latent $z_t$ and fed into the model.

\paragraph{Classifier-Free Guidance (CFG)} The CFG~\cite{ho2022classifier} is a technique employed in diffusion models to balance image quality and prompt fidelity during the generation process. It operates by mixing conditional sampling $\epsilon_\theta(z_t,t,\mathbf{e})$, performed using the desired prompt embedding $e$, with unconditional sampling $\epsilon_\theta(z_t, t, \mathbf{e}_n)$, performed using the null text prompt embedding $e_n$, to control the degree of conditioning as follows:
\begin{equation}
\label{Eq: CFG}
    \hat\epsilon_\theta(\tilde z_t,t,\mathbf{e})=(1+w)\cdot \epsilon_\theta(\tilde z_t,t,\mathbf{e})-w\cdot \epsilon_\theta(\tilde z_t,t,\mathbf{e}_n),
\end{equation}
where $w$ is the CFG scale, which is typically user-adjustable. As $w$  increases, the influence of unconditional sampling diminishes while that of conditional sampling strengthens, leading to a more faithful reflection of the given prompt.

\subsection{Adversarial Attack on Inpainting Model}
\label{Subsec: adversarial attack on inpainting model}
Adversarial attacks~\cite{goodfellow2014explaining,kurakin2018adversarial,madry2017towards,szegedy2013intriguing,biggio2013evasion} deceive neural network models by introducing small imperceptible perturbations in input data. A widely used method, the Projected Gradient Descent (PGD)~\cite{madry2017towards} attack, iteratively maximizes the loss function while constraining the perturbations within a predefined boundary, as follows:
\begin{equation}
\label{Eq: PGD attack}
    x_{i+1}=\text{Proj}_{\mathcal{B}_{\epsilon}}(x_i + \alpha \cdot \text{sign}(\nabla_xL(f_\theta(x_i),y))),
\end{equation}
where $x_i$ is image with current adversarial noise, $y$ is ground truth label, $f_\theta$ is model, $L$ is loss function, $\alpha$ is step size, $\text{Proj}_{\mathcal{B}_{\epsilon}}$ is a projective function with $\epsilon$ budget constraints.

To prevent malicious users from freely inpainting parts of an image, previous studies~\cite{son2024disrupting,salman2023raising,choi2024diffusionguard, jeon2025advpaint} have explored methods for injecting adversarial noise into images. This noise disrupts the inpainting process, hindering the diffusion model from effectively manipulating the perturbed images, thereby offering protection against unauthorized and harmful usage. These methods generate noise to suppress inpainting by either maximizing or minimizing variations of the following general objective function which is simply expressed as:
\begin{equation}
\label{Eq: PGD on inpainting model}
    \begin{array}{cc}
         L_{\text{adv}}=||y_{\text{target}} - f_\theta(\tilde z_t^\prime,t, \mathbf{e})||_p ,\quad \text{s.t. } ||\delta||_\infty \le \epsilon ,
    \end{array}
\end{equation}
where $\tilde z_t^\prime$ is $\text{concat}(z_t,\mathbf{M}^\prime,\mathcal{E}((x+\delta) \otimes \mathbf{M}))$,  $\delta$ is adversarial noise, $\epsilon$ is noise budget, $f_\theta$ is a part of the noise prediction model, $y_{\text{target}}$ is the target defined by each method to get closer to or move away from, and $p$ represents the $p$-norm.

\section{Method}
\label{Sec: Method}

\begin{figure*}[ht]
  \centering
  \includegraphics[width=0.8\linewidth]{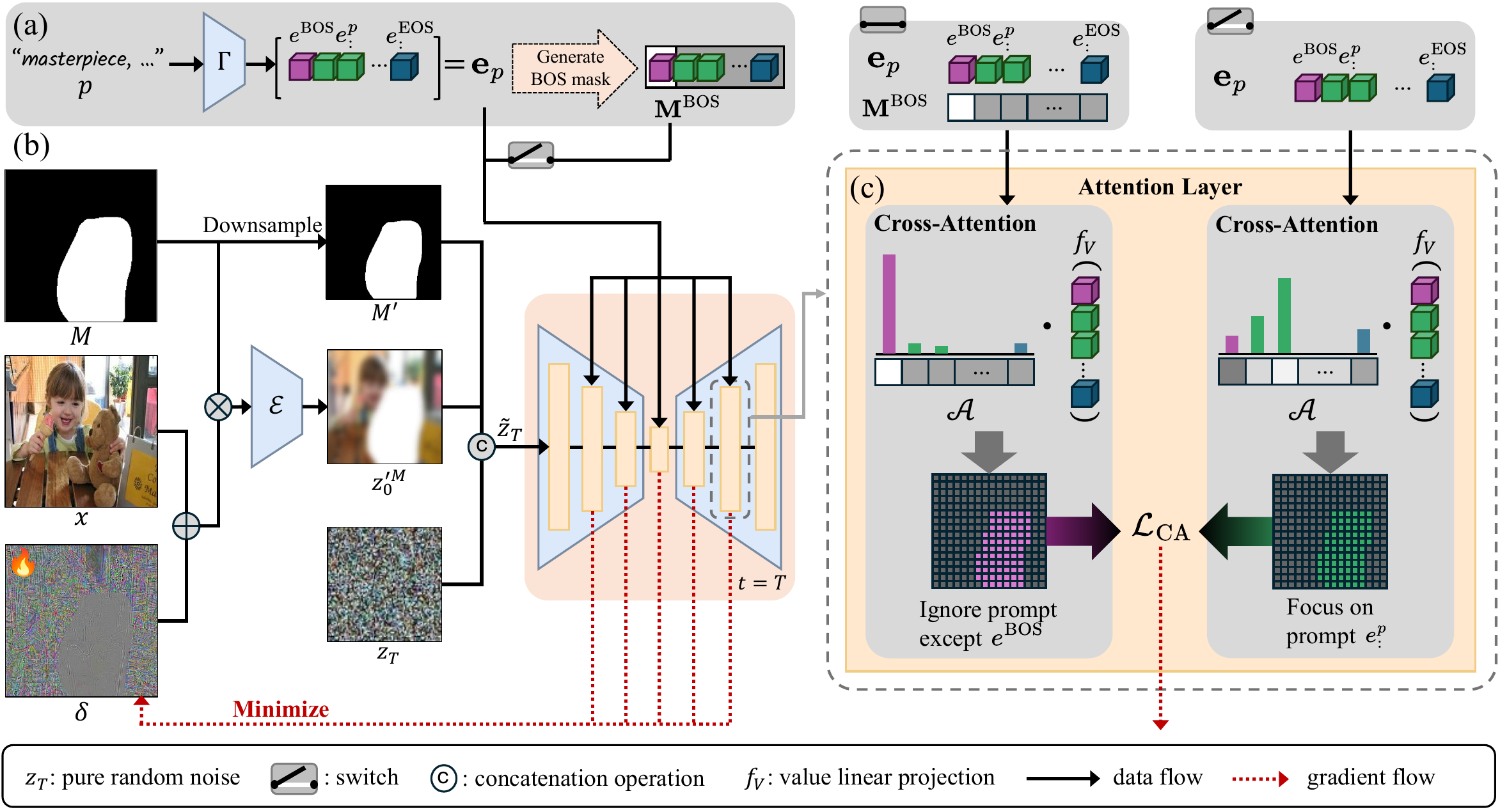}
  \caption{Overview of our method, \SystemName. In the prompt preprocessing stage (a), the prompt is tokenized and embedded, and a BOS mask is generated. In the data input stage (b), two versions of the input—with and without the BOS mask—are batched and directly fed into each attention layer. In the attention layer (c), cross-attention is computed in distinct ways depending on whether the BOS mask is applied. Adversarial noise is then updated to minimize the difference between the two outcomes.}
  \Description{The overview of our method.}
  \label{Fig: method figure}
\end{figure*}

\subsection{Problem Statement}
\label{Subsec: Problem statement}

We consider a realistic scenario in which a defender preemptively injects adversarial noise into an image to protect against unauthorized manipulation. Later, a malicious user attempts to alter this protected image using a publicly available Stable Diffusion inpainting model. A fundamental challenge arises: the defender has no knowledge of the prompt that the attacker will use, as the attack occurs after the protection has been applied. This renders any defense mechanism that relies on prompt-specific information fundamentally insufficient.

Previous studies~\cite{son2024disrupting,salman2023raising,choi2024diffusionguard, jeon2025advpaint} have attempted to address this challenge by simply treating the prompt embedding as either a null text embedding or as the text embedding that best describes the given image~\cite{gal2022image}, and then maximizing image-level inconsistencies. While these methods may partially disrupt the unconditional sampling process, they are ineffective against conditional sampling, which directly aligns image generation with the semantic content of the prompt. Furthermore, increasing the CFG scale reinforces this alignment, allowing the model to disregard perturbations. Consequently, even heavily perturbed images can produce outputs that remain faithful to malicious prompts under high CFG settings.

PromptFlare addresses this limitation by explicitly targeting the interaction between the prompt and the image via the cross-attention mechanism. We propose two novel strategies: (1) isolating a shared token invariant across all prompts and enforcing attention solely on it, and (2) ensuring this token is semantically meaningless to prevent the model from accessing prompt-relevant information. This prompt-generalized design ensures that the protection remains effective regardless of the attacker’s input prompt or CFG scale. 

In Sec \ref{Subsec: shared component}, we examine the process of converting prompts into text embeddings using a text encoder and identify tokens that are universally present across all prompts but carry no meaningful information. In Sec \ref{Subsec: adversarial attack on CA}, we explore the mechanism of cross-attention and define an adversarial objective that directly disrupts the cross-attention between the image and the prompt. In Sec \ref{Subsec: quality tag prompt}, we discuss the selection of the quality tag prompt used for actual computing and introduce the final objective function.

\subsection{Shared Tokens of Prompts}
\label{Subsec: shared component}

To input a prompt $p$ into the inpainting model, the prompt is first converted into a prompt embedding $\mathbf{e}$ using a text encoder $\Gamma$. Specifically, the prompt is tokenized into $|p|$ tokens, prepended with a Beginning-of-Sequence (BOS) token, and appended with $N - |p| - 1$ End-of-Sequence (EOS) tokens, resulting in a total length of $N$. The resulting prompt embedding $\mathbf{e}$ is then expressed as follows:

\begin{equation}
\label{Eq: prompt to embedding}
    \mathbf{e}= \Gamma(p)=[e^{\text{BOS}}, e^{p}_0, ..., e^{p}_{|p|-1}, e^{\text{EOS}}_{0}, ...,e^{\text{EOS}}_{N-|p|-2}]\in \mathbb{R}^{N \times K}.
\end{equation}
Fig. \ref{Fig: method figure} (a) illustrates our prompt preparation process. If $N-2<|p|$, tokens beyond the $(N-2)\text{-th}$ position are truncated, and a single EOS token is appended at the end. For the Stable Diffusion model, $\Gamma$ is a pre-trained CLIP~\cite{radford2021learning} text encoder, with $N=77$ and $K=768$.

Examining Eq. \ref{Eq: prompt to embedding}, we observed that for any given prompt $p$, the two token embeddings $e^{\text{BOS}}$ and $e^{\text{EOS}}_{N-|p|-2}$ are always present. As both are special tokens, they may initially appear to have no meaningful information. However, due to the design and operational principles of the CLIP text encoder, it has been revealed that later tokens in the sequence incorporate information from the preceding tokens within their embeddings~\cite{radford2021learning, li2024get}. 
Consequently, the embedding $e^{\text{EOS}}_{N-|p|-2}$ encapsulates comprehensive information about the entire prompt $p$. Thus, even if some tokens are ignored, the model can still effectively acquire the information in $p$  through $e^{\text{EOS}}_{N-|p|-2}$.
In contrast, since $e^{\text{BOS}}$ appears at the beginning of the sequence, it does not encode any information about the prompt $p$. Based on this observation, we select $e^{\text{BOS}}$ and construct a mask $\mathbf{M}^{\text{BOS}} \in \mathbb{R}^{1 \times N}$ that exclusively targets this token, as defined below:

\begin{equation}
\label{Eq: BOS mask}
    M^{\text{BOS}}_i=
    \begin{cases}
        1, & \text{if } e_i = e^{\text{BOS}}\\
        0, & \text{otherwise}
    \end{cases}.
\end{equation}

We analyze the selection of alternative shared tokens, specifically $e^{\text{EOS}}_{N-|p|-2}$ instead of the default choice $e^{\text{BOS}}$, through an ablation study presented in Sec. \ref{Subsec: ablation study}.

As described in Sec. \ref{Subsec: adversarial attack on inpainting model}, the input to the noise prediction model, denoted as $\tilde{z}_T^\prime$, is constructed from the image mask $M$, the original image $x$, and the adversarial noise $\delta$.
To perform the adversarial attack, we feed into each attention layer of the noise prediction model a batch consisting of two inputs: the prompt embedding $\mathbf{e}$ with the BOS mask $\mathbf{M}^{\text{BOS}}$, and the same embedding without the mask. Fig. \ref{Fig: method figure} (b) illustrates our input construction process clearly.

\subsection{Adversarial Attack on Cross-Attention}
\label{Subsec: adversarial attack on CA}

The noise prediction model in Stable Diffusion follows a U-Net architecture and comprises $L$ attention blocks, each of which includes a cross-attention mechanism. When computing $\epsilon_\theta (\tilde{z}_t,t, \mathbf{e})$, the prompt embedding $\mathbf{e}$ influence exclusively through the cross-attention. Consequently, we aim to disrupt this cross-attention mechanism. The cross-attention optionally takes a prompt mask $\mathbf{M}_{\text{p}}$ of length $N$, corresponding to the dimensionality of the prompt embedding. The attention map $\mathcal{A}$ is then computed as follows:
\begin{equation}
\label{Eq: Attention map}
\mathcal{A}(\phi(\tilde z_t), \mathbf{e},\mathbf{M}_{{p}})=\text{softmax} \left( {\frac{QK^T}{\sqrt{d}}}+(\mathbf{M}_p \times c) \right),
\end{equation}
where $\phi(\tilde{z}_t)$ is the output feature of previous layer before cross-attention, $Q$ is query extracted from the $\phi(\tilde{z}_t)$, $K$ is the key extracted from the prompt embedding $\mathbf{e}$, $d$ is the dimension of $Q$ and $c$ is a large positive constant. Eq. \ref{Eq: Attention map} means that the attention map $\mathcal{A}$ exhibits a substantially increased probability of attending exclusively to the masked tokens across all regions. In the absence of a mask, the probability increases that the model attends to tokens most suitable for each spatial region, thereby extracting prompt information as intended by the pre-trained model. Fig. \ref{Fig: method figure} (c) illustrates this mechanism of cross-attention using attention map. The corresponding cross-attention output is computed as follows:
\begin{equation}
\label{Eq: cross attention softmax}
    \text{CA}(\phi(\tilde z_t), \mathbf{e},\mathbf{M}_{{p}})= \mathcal{A}(\phi(\tilde z_t), \mathbf{e},\mathbf{M}_{{p}})V,
\end{equation}
where $V$ is value extracted from the prompt embedding $\mathbf{e}$.

By employing the previously defined prompt embedding mask $\mathbf{M}^{\text{BOS}}$ as $\mathbf{M}_{\text{p}}$, we aim to achieve the following two effects: 1) Prompt-generalized behavior: As discussed in Sec. \ref{Subsec: shared component}, the token embedding $e^{\text{BOS}}$ is always present, regardless of the given prompt. By focusing exclusively on $e^{\text{BOS}}$ while disregarding the remainder of $\mathbf{e}$, our approach ensures consistent influence on both conditional and unconditional sampling processes. 2) Disruption of cross-attention: As previously discussed, the attention map becomes heavily biased toward attending exclusively to the masked token $e^{\text{BOS}}$ across all regions, including the inpainting area. Since $e^{\text{BOS}}$ carries no information pertaining to the prompt $p$, the cross-attention mechanism fails to incorporate any semantic content from the original prompt into the output features. 

Consequently, our approach consistently induces the disruption of the cross-attention mechanism during the sampling process. 
Given these desirable properties, utilizing $\mathbf{M}^{\text{BOS}}$ as the prompt mask in the cross-attention mechanism provides an effective and straightforward solution to the limitations observed in previous approaches
, as discussed in Sec. \ref{Subsec: Problem statement}. Therefore, we define the target as the cross-attention trajectory computed with $\mathbf{M}^{\text{BOS}}$ and construct a loss term by comparing it to the cross-attention output obtained without $\mathbf{M}^{\text{BOS}}$:
\begin{equation}
\label{Eq: almost final loss}
    L_{\text{CA}}=|| \text{CA} (\phi(\tilde z_t),\mathbf{e},\mathbf{M}^{\text{BOS}}) \otimes \mathbf{M}^\prime -\text{CA} (\phi(\tilde z_t),\mathbf{e}) \otimes \mathbf{M}^\prime||_2 .
\end{equation}
Here, since we are specifically concerned with the suppression occurring in the inpainting region, the comparison is made only over the region indicated by the image latent mask $\mathbf{M}^\prime$.

\begin{figure*}[th]
  \centering
  \includegraphics[width=0.9\linewidth]{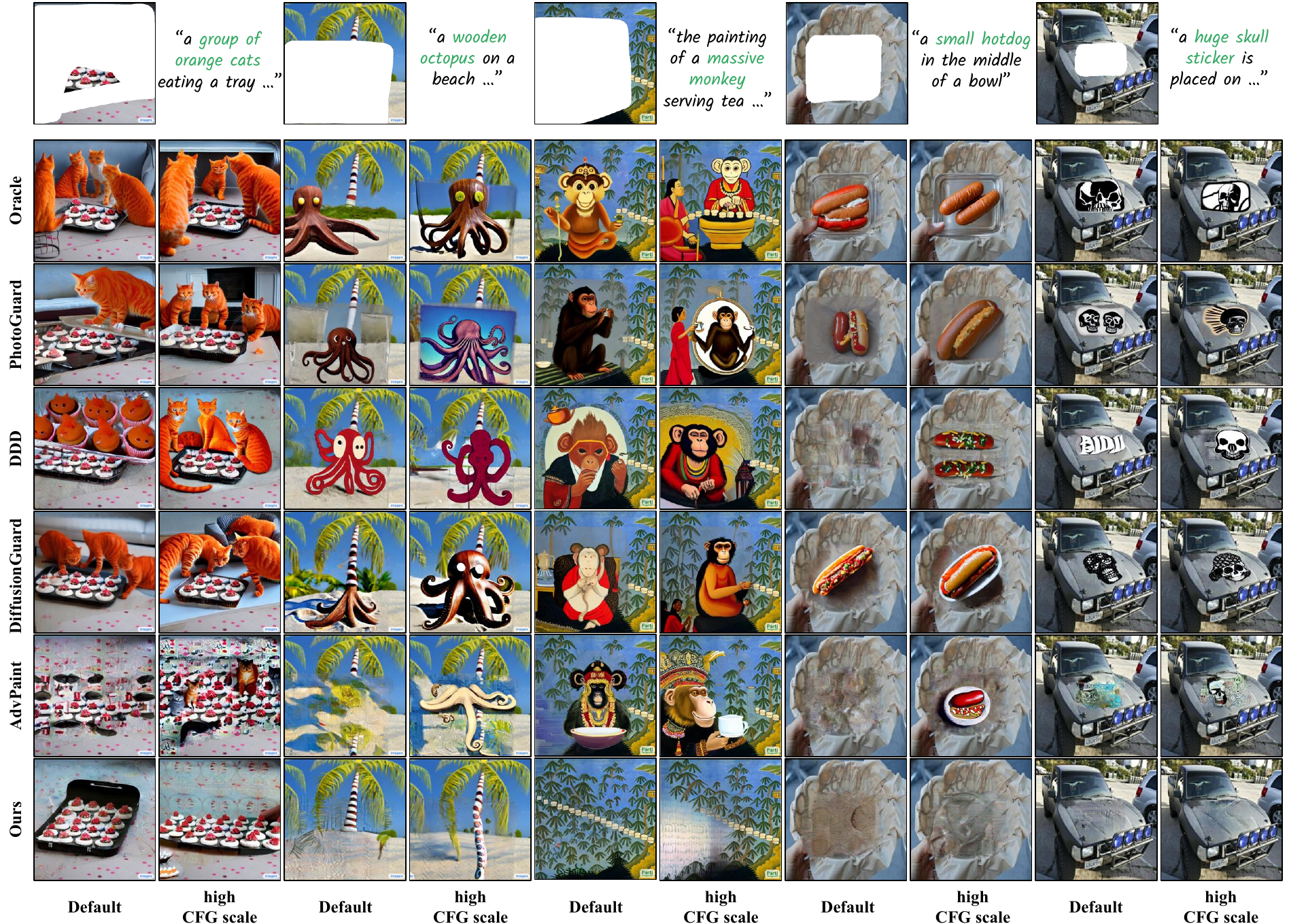}
  \caption{Qualitative evaluation of various methods across different CFG scales. The Default is set to 7.5, and the high CFG scale ranges from 12.5 to 15.0. The illustrated results show that previous SoTA methods, despite occasional success at low CFG scales, tend to follow the prompt as the scale increases. Our method consistently protect the image regardless of CFG scale.}
  \Description{Figure of experimental results across various CFG scale.}
  \label{Fig: cfg scale}
\end{figure*}

\subsection{Quality Tag Prompt}
\label{Subsec: quality tag prompt}

To enhance the effectiveness of the adversarial attack, we adopt a quality tag prompt as the base prompt for computing adversarial noise. This prompt comprises semantically neutral yet highly influential tokens, such as ``masterpiece'' and ``best quality'', which are widely recognized for improving the visual fidelity of diffusion-based image generation models when appended to a prompt. Although these tokens do not describe specific objects or spatial attributes, their consistent empirical effect suggests that they bias the generative process toward aesthetically enhanced outputs.

By using the quality tag prompt embedding, we deliberately introduce a strong generative prior into the diffusion process—one that the model is predisposed to follow in order to enhance image quality. Under this condition, our objective is to constrain the model to attend exclusively to the BOS token $e^{\text{BOS}}$. This leads the adversarial attack to generate noise that more effectively drives the model to disregard the prompt compared to using a null prompt alone. The results of the ablation study in Sec. \ref{Subsec: ablation study} verify the effectiveness of this approach.

The final objective function, incorporating all of our proposed methods, is formulated as follows:
\begin{equation}
\label{Eq: final loss}
    \argmin_{||\delta||_\infty\le\epsilon} \mathbb{E}_{l} [L_\text{CA}^l],
\end{equation}
where $l$ denotes the $l\text{-th}$ layer in the noise prediction model. Empirically, $l$ is set to include only the attention layers excluding the outermost one, as the latter primarily captures overly fine-grained information, which is less relevant for the objective. For a detailed ablation study on the selection of attention layers, please refer to Appendix C.1.


\section{Experiments}
\label{section: experiments}
\subsection{Experimental Details}
\paragraph{Dataset} To evaluate the effectiveness of our method, we conducted comprehensive experiments under various conditions using the EditBench~\cite{wang2023imagen} dataset. The EditBench dataset comprises a total of 240 images, including 120 real images and 120 generated images, each paired with a corresponding mask image. This dataset is significantly larger than the private evaluation datasets used in previous methods~\cite{son2024disrupting, choi2024diffusionguard}, which contain only 13 and 42 images, respectively. Furthermore, EditBench provides distinct prompts for the entire image and the masked region, enabling separate evaluations for each. For more details on the dataset, see the Appendix A.

\paragraph{Baselines} For comparison with our method, we selected PhotoGuard~\cite{salman2023raising}, DDD~\cite{son2024disrupting}, DiffusionGuard~\cite{choi2024diffusionguard} and AdvPaint~\cite{jeon2025advpaint} as baselines, all of which apply adversarial attacks to inpainting models. In addition, we include Oracle as a reference baseline, which refers to the result of inpainting obtained from the original image without any applied protection. Each baseline was implemented using the best hyperparameters reported in its respective paper. However, to ensure a fair comparison, we set a consistent noise budget across all methods in our primary experiments, while also exploring different noise budget values in the Appendix C.6.

\paragraph{Evaluation Metrics} We employed two types of metrics: 

1) Reference-free metrics. CLIP Score~\cite{hessel2021clipscore}, Aesthetic Score~\cite{huang2024predicting}, and PickScore~\cite{kirstain2023pick} were selected to quantitatively assess how well the inpainted images align with the given prompts and to evaluate their aesthetic quality. Specifically, the CLIP Score measures semantic consistency between image and text embeddings, while the Aesthetic Score evaluates the visual attractiveness of generated images. PickScore, originally designed to select the best-matching image among candidates using a Softmax operation, was utilized differently in our work. We excluded the Softmax step and directly used its raw scoring output to gauge alignment quality more precisely. Lower scores of three metrics indicate better suppression of the prompt influence. Evaluations were performed separately for the entire inpainting image (`all') and for the masked region (`mask').

2) Reference-based metrics. To ensure experimental consistency with previous works, we report results using traditional metrics that rely on reference images. LPIPS~\cite{zhang2018unreasonable}, SSIM~\cite{wang2004image}, and PSNR evaluate the visual differences between the inpainting results using the protected image with each method and the Oracle. LPIPS evaluates perceptual similarity using CNN-extracted features.
SSIM measures structural consistency by comparing luminance, contrast, and structural information.
PSNR measures pixel-level differences, indicating the degree of distortion or noise introduced. Higher LPIPS and lower SSIM/PSNR scores suggest greater perceptual differences with Oracle. These metrics are limited by the assumption that the reference image constitutes the sole ground truth.

\paragraph{Experimental Setup} For the implementation of our method, we utilized an NVIDIA A5000 GPU with 24GB of memory. Unless otherwise specified, our default settings for adversarial noise employed the Stable Diffusion inpainting model from RunwayML. The specific parameters were as follows: PGD epsilon budget of 12/255, a step size of 2/255, and one-step gradient averaging repeated over 400 iterations. For the inpainting process, we set the CFG scale to 7.5, the inference steps to 50, and the strength to 1.0.

\subsection{Experimental Results}

\paragraph{Qualitative Results} As shown in Default columns of Fig. \ref{Fig: cfg scale}, while some baseline methods faithfully adhere to the prompt by inpainting the key object into the image, our approach achieves effective generation suppression. Importantly, rather than relying on image-level inconsistencies, our method aims to eliminate the prompt's influence on the image entirely. Consequently, the resulting inpainted regions more closely
resemble outputs generated using only the image context without prompts, as opposed to displaying unnatural patterns or color shifts. 

\begin{table*}[th]
    \centering
    \caption{Quantitative comparison of various suppression methods using reference-free and reference-based metrics on the EditBench dataset. ``all'' uses the full image and prompt (including background), while ``mask'' uses only the masked region and its prompt (excluding background). Best scores are highlighted in bold, and the second-best results are underlined. Our method achieves state-of-the-art performance, with especially large gains in CLIP Score.}
    \label{Tab: main result}
    \begin{tabular}{c c c c c c c c c c}
    \toprule
    \multirow{2}{*}{Methods}
        & \multicolumn{2}{c}{CLIP Score $\downarrow$}
        & \multicolumn{2}{c}{Aesthetic $\downarrow$}
        & \multicolumn{2}{c}{PickScore $\downarrow$}
        & \multirow{2}{*}{LPIPS $\uparrow$}
        & \multirow{2}{*}{SSIM $\downarrow$}
        & \multirow{2}{*}{PSNR $\downarrow$} \\
    \cmidrule(lr){2-3}
    \cmidrule(lr){4-5}
    \cmidrule(lr){6-7}
    & \textit{all} & \textit{mask}
    & \textit{all} & \textit{mask}
    & \textit{all} & \textit{mask}
    &  &  &  \\
    \midrule
    Oracle
        & 32.0412 & 26.6546
        & 4.7981  & 4.3493
        & 20.7639 & 19.6893
        & --      & --      & --      \\
    PhotoGuard~\cite{salman2023raising}
        & 31.7589 & 26.3759
        & 4.5361  & 4.2780
        & 20.4062 & 19.6097
        & 0.2443  & 0.6947  & 14.8245 \\
    DDD~\cite{son2024disrupting}
        & 30.9199 & 25.2935
        & 4.5625  & 4.5685
        & 20.2364 & 19.3621
        & 0.2449  & 0.6969  & \textbf{14.2616} \\
    DiffusionGuard~\cite{choi2024diffusionguard}
        & 30.6709 & 25.1593
        & 4.6233  & 4.2097
        & 20.3921 & 19.3010
        & 0.2547  & 0.7016  & 14.5926 \\
    AdvPaint~\cite{jeon2025advpaint}
        & \underline{28.2006} & \underline{22.4697}
        & \textbf{4.2765}  & \underline{4.1440}
        & \textbf{19.4995} & \underline{18.7295}
        & \textbf{0.2728}  & \textbf{0.6802}  & \underline{14.2763} \\
    \rowcolor{backcolour}\SystemName~(\textit{\textbf{ours}})
        & \textbf{27.3171} & \textbf{21.9557}
        & \underline{4.4569}  & \textbf{4.0916}
        & \underline{19.5498} & \textbf{18.6491}
        & \underline{0.2570}  & \underline{0.6894}  & 14.9311 \\
    \bottomrule
    \end{tabular}
\end{table*}
\begin{figure*}[th]
  \centering
  \includegraphics[width=0.9\linewidth]{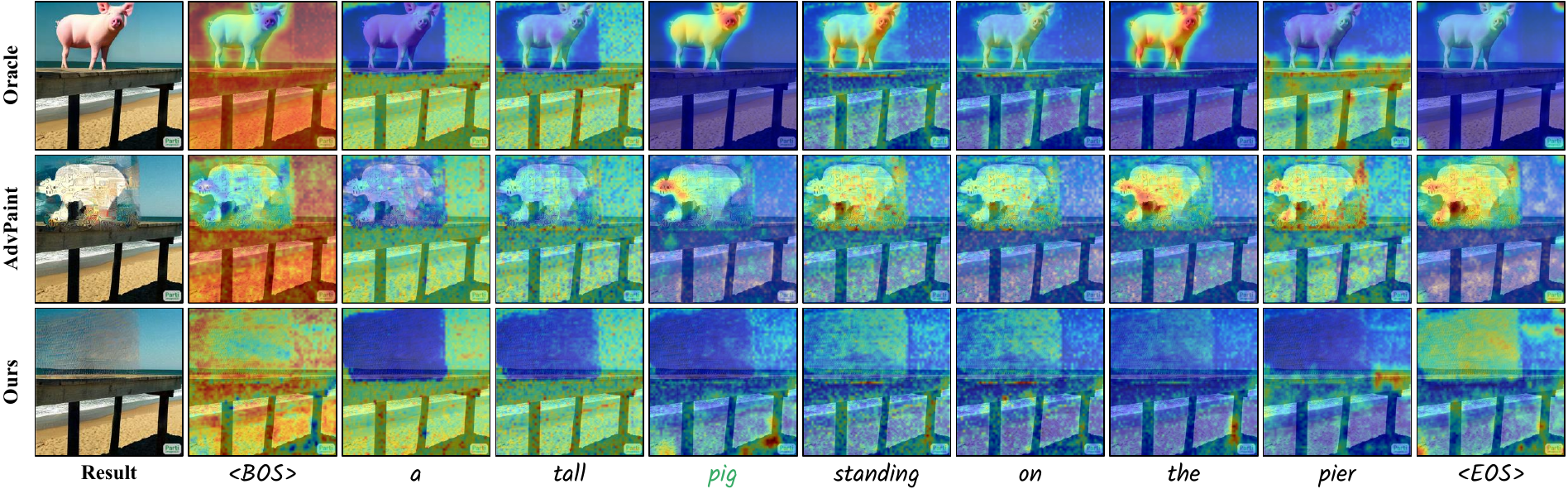}
  \caption{We visualize cross-attention maps for each token. In case of Oracle and AdvPaint, the green token and EOS token show strong activation over the masked region. In contrast, with our method, the green token has little to no influence, and the BOS token becomes the primary contributor.}
  \Description{attention maps activation figure.}
  \label{Fig: attention map}
\end{figure*}

\paragraph{Quantitative Results} As shown in Tab. \ref{Tab: main result}, our method significantly outperforms other baselines in CLIP-based metrics. In particular, our proposed method outperforms the previous SoTA method, AdvPaint, by more than 0.88 points in CLIP Score. 
Our method achieves SoTA on Aesthetic Score and PickScore within the inpainted regions. This indicates that the ~\SystemName~ effectively disrupts the alignment between the generated image and the prompt. In reference-based metrics, our method ranks near the second-best but exhibits relatively low performance in PSNR. This is attributed to inherent limitations in the design of these metrics. They inherently assume that the reference image represents the sole ground truth, which limits their ability to capture failure cases where the output closely follows a malicious prompt while deviating from the Oracle. This limitation is especially pronounced in PSNR, which performs strict pixel-wise comparisons and is therefore sensitive to minor variations such as slight noise, changes in object size, or positional shifts. The main goal of our method is not to induce image-level noise or prevent oracle reconstruction, but to neutralize the influence of malicious prompts. Therefore, we places greater emphasis on reference-free metrics that more directly reflect the goals of prompt-level defense.

\subsection{Comparison on CFG Scale}
In this section, we compare the inpainting results based on the CFG scale, a tunable parameter available to malicious users. As shown in Fig. \ref{Fig: cfg scale}, existing methods occasionally achieve generation suppression at low CFG scales. However, such suppression is often unstable, with the models frequently reflecting the prompt content. Even in cases when suppression appears successful, simply increasing the CFG scale with the same input typically leads to the reemergence of the intended generation. This observation supports our claim that prior methods lack robustness in handling conditional sampling. In contrast, our approach demonstrates stable and consistent generation suppression across all CFG scales.

\begin{figure}[t!]
  \centering
  \includegraphics[width=0.9\linewidth]{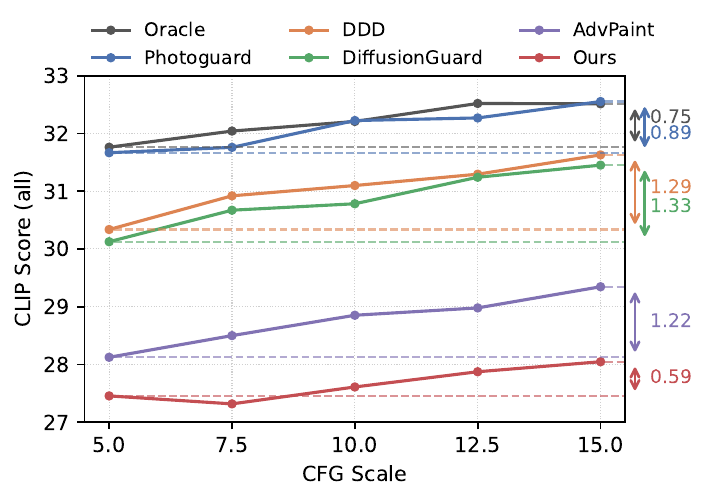}
  \caption{Comparison of CLIP Score (all) across different CFG scales. Across all CFG scales, our method yields lower values than the previous SoTA at the minimum scale setting. It also shows the smallest score gap between CFG 5.0 and 15.0. 
  }
  \Description{The Graph of change of CLIP Score across various CFG scale on each methods.}
  \label{Fig: cfg scale graph}
\end{figure}

Figire \ref{Fig: cfg scale graph} illustrates the variation in CLIP Scores for each method as the CFG scale changes. Across all scales, our method consistently outperforms all baselines. Notably, even at the highest CFG scale, our method achieves a lower score than the previous SoTA method at its minimum CFG setting, demonstrating superior robustness under maximal guidance. While previous methods fail to maintain suppression at higher CFG scales, our method demonstrates robust protection, as evidenced by a minimal increase in CLIP Score even at high CFG scales. This clearly indicates that our approach effectively mitigates prompt conditioning. Moreover, when comparing CFG scales of 5.0 and 15.0, the increase in CLIP Score for our method across CFG scales is less than half that of previous methods, indicating superior robustness to prompt influence. These results indicate that our method effectively suppresses prompt influence, thereby robustly addressing the challenge of conditional sampling. More detailed results regarding the impact of CFG scale are provided in Appendix C.2.

\begin{table*}[th]
    \centering
    \caption{Quantitative comparison from the ablation study. ``Use EOS token'' sets the mask on $e^{\text{EOS}}_{N - |p| - 2}$ instead of $e^{\text{BOS}}$ to compute adversarial noise. The result is nearly identical to the unprotected case. ``Use null prompt'' replaces the quality tag prompt embedding with the null prompt. This case still provides protection, but less effectively than \SystemName.}
    \label{Tab: ablation result}
    \begin{tabular}{c c c c c c c c c c}
    \toprule
    \multirow{2}{*}{Methods}
        & \multicolumn{2}{c}{CLIP Score $\downarrow$}
        & \multicolumn{2}{c}{Aesthetic $\downarrow$}
        & \multicolumn{2}{c}{PickScore $\downarrow$}
        & \multirow{2}{*}{LPIPS $\uparrow$}
        & \multirow{2}{*}{SSIM $\downarrow$}
        & \multirow{2}{*}{PSNR $\downarrow$} \\
    \cmidrule(lr){2-3}
    \cmidrule(lr){4-5}
    \cmidrule(lr){6-7}
    & \textit{all} & \textit{mask}
    & \textit{all} & \textit{mask}
    & \textit{all} & \textit{mask}
    &  &  &  \\
    \midrule
    Oracle & 32.0412 & 26.6546 & 4.7981 & 4.3493 & 20.7639 & 19.6893 & -- & -- & -- \\
    \midrule
    \rowcolor{backcolour}\SystemName~\textbf{(\textit{ours})} & \textbf{27.3171} & \textbf{21.9557} & \textbf{4.4569} & \textbf{4.0916} & \textbf{19.5498} & \textbf{18.6491} & \textbf{0.2570} & \textbf{0.6894} & 14.9311 \\
    + Use EOS token & 32.4044 &  26.7319 & 4.6740 & 4.3366 & 20.6020 & 19.6336 & 0.2577 & 0.6918 & \textbf{13.9021} \\
    + Use null prompt & 28.9199 & 23.3096 & 4.4758 & 4.1012 & 19.8934 & 18.9149 & 0.2559 & 0.6914 & 14.7674 \\
    \bottomrule
    \end{tabular}
\end{table*}

\subsection{Visualize Attention Map}
\label{Subsec: attention map}

To empirically verify that our method disrupts the cross-attention and prevents the model from focusing on the prompt, we visualize the token-wise attention maps during the inpainting process using DAAM~\cite{tang2023daam}. As illustrated in Fig. \ref{Fig: attention map}, the Oracle case shows that the green token, which represents the core object intended for inpainting, exerts strong influence over the inpainting region. Meanwhile, the BOS token mainly attends to areas outside the inpainting region, the image context. A similar trend is observed in the previous SoTA, AdvPaint. Although it may initially appear to suppress generation by producing unnatural white object and distorted color patterns, attention map analysis reveals substantial involvement of the green token and the EOS token within the inpainted region. This observation suggests that increasing the CFG scale can lead to more faithful align to the prompt, which may undermine the effectiveness of the defense. However, in our method, only the BOS token exhibits influence within the inpainting region. All other tokens, especially ``pig'', show no meaningful contribution. This analysis provides empirical evidence that our method successfully suppresses the influence of the prompt during generation. Token-wise attention visualizations for the inpainted images generated by different methods are provided in the Appendix C.3.

\subsection{Ablation Study}
\label{Subsec: ablation study}
To empirically assess the validity of our design choices, we performed an ablation study examining the effects of key components in our method. As mentioned in Tab. \ref{Tab: ablation result}, the ``null prompt'' condition involves substituting the quality tag prompt with a null prompt, and the ``EOS token'' setting applies the attention mask to the EOS token rather than the BOS token. The null prompt variant yields inferior performance relative to the quality tag prompt in nearly all metrics, with the exception of PSNR. Furthermore, when focusing on the EOS token, no protection is observed, supporting previous research findings that the EOS token aggregates information from preceding tokens.

We conducted further experiments on unseen masks, model transferability, and comparisons across various noise budgets, inference steps, strength parameters, and robustness conditions. Our method achieved best performance in nearly all tasks. Due to space constraints, detailed results of these experiments are provided in the Appendix C.

\subsection{Efficiency Analysis}

\begin{figure}[t!]
  \centering
  \includegraphics[width=\linewidth]{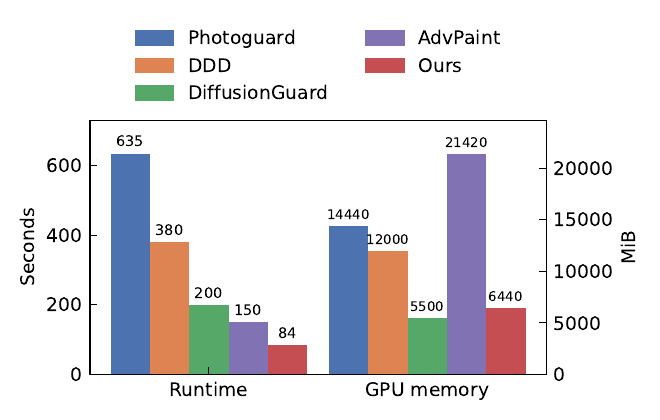}
  \caption{Comparison of runtime and GPU memory consumption. Our method achieves significantly lower runtime, running at least twice as fast as baseline methods, while also maintaining the lowest GPU memory usage.}
  \Description{The Graph of runtime and gpu memory across various CFG scale on each methods.}
  \label{Fig: efficiency result}
\end{figure}

Figure. \ref{Fig: efficiency result} illustrates the efficiency of our method. It achieves generation suppression at speeds that are about twice as fast and up to seven and a half times faster than the baseline methods, while also demonstrating some of the lowest GPU memory consumption. This efficiency results from the use of an objective function that is carefully aligned with the intended goal, enabling effective suppression with minimal computational overhead. As a consequence, the method increases accessibility by providing more users with the ability to protect their images efficiently.

\section{Conclusion}

In this paper, we presented \SystemName, a novel and effective adversarial defense mechanism for diffusion-based inpainting models. By exploiting the invariant properties of prompt embeddings, \SystemName~ construct an adversarial noise that functions as a cross-attention decoy. When injected into the image, it systematically disrupts the cross-attention process and prevents the model from faithfully reflecting malicious prompts. Through comprehensive experimental analysis, we demonstrated that \SystemName~ achieves SoTA performance across most metrics while maintaining high computational efficiency and low resource consumption. These results not only underscore the robustness of our approach in real-world scenarios but also pave the way for further research into prompt-generalized defenses.

\begin{acks}
The authors would thank anonymous reviewers. Simon S. Woo is the corresponding author. This work was partly supported by Institute for Information \& communication Technology Planning \& evaluation (IITP) grants funded by the Korean government MSIT:
(RS-2022-II221199, RS-2022-II220688, RS-2019-II190421, RS-2023-00230337, RS-2024-00356293, RS-2024-00437849, RS-2021-II212068,  RS-2025-02304983, and  RS-2025-02263841).
\end{acks}

\bibliographystyle{ACM-Reference-Format}
\balance
\bibliography{reference}

\appendix
\clearpage
\setcounter{page}{1}
\setcounter{section}{0}
\setcounter{figure}{0}
\setcounter{equation}{0}
\setcounter{table}{0}

\large\textbf{Supplementary Material} \\ 

\section{Dataset}
\label{Sec: dataset}

\begin{figure}[th]
  \centering
  \includegraphics[width=\linewidth]{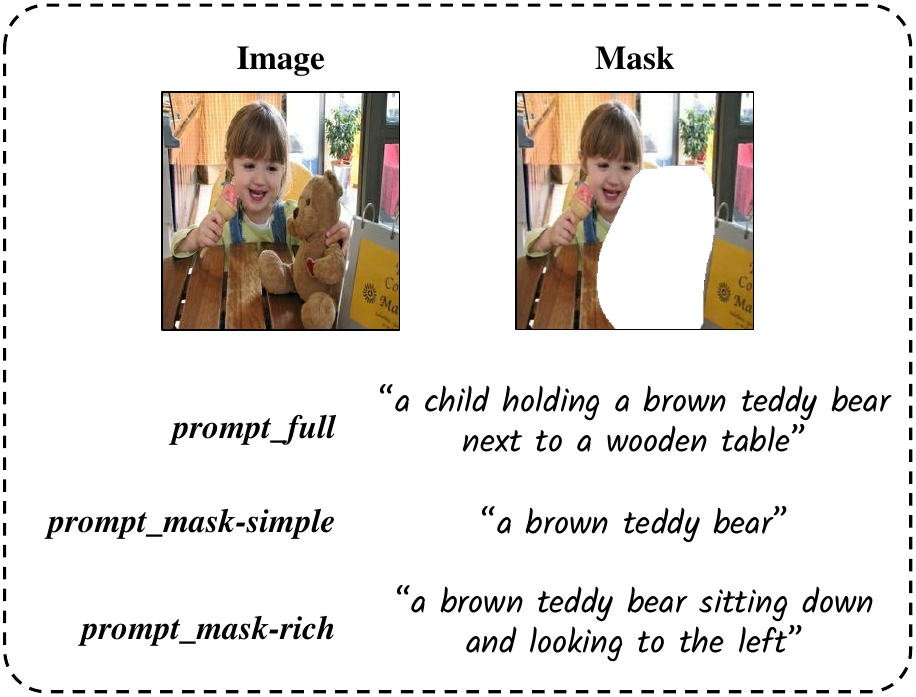}
  \caption{An example image pair from the EditBench~\cite{wang2023imagen} dataset. Each pair includes the original image, a mask indicating the inpainting region, and distinct types of prompts associated with the image.}
  \Description{A woman and a girl in white dresses sit in an open car.}
  \label{AFig: dataset}
\end{figure}

The EditBench~\cite{wang2023imagen} dataset consists of 240 image pairs, including 120 real images and 120 generated images. As shown in Fig. \ref{AFig: dataset}, each pair comprises an original image, a mask specifying the region to be inpainted, a full-scene description (\textit{prompt$\_$full}), a simplified prompt describing only the core object in the masked region (prompt$\_$mask-simple), and a more detailed description of the masked region (\textit{prompt$\_$mask-rich}).  Notably, the mask is not strictly binary. In our experiments, \textit{prompt\_full} is used as the input for inpainting. We evaluate alignment in two ways: (1) between \textit{prompt\_full} and the entire inpainted image, and (2) between \textit{prompt\_mask-simple} and the inpainted masked region. This protocol follows the findings of the EditBench authors, who observed that inpainting outputs generally align more closely with simpler prompts~\cite{wang2023imagen}.  Unlike prior methods that rely on smaller, private datasets, our use of this larger scale, publicly benchmark, which enables separate evaluation of global and localized prompt alignment allows for a more comprehensive and equitable assessment of our method.

\section{Noise Visualization}
\label{Sec: noise visualization}

\begin{figure*}[th]
  \centering
  \includegraphics[width=\linewidth]{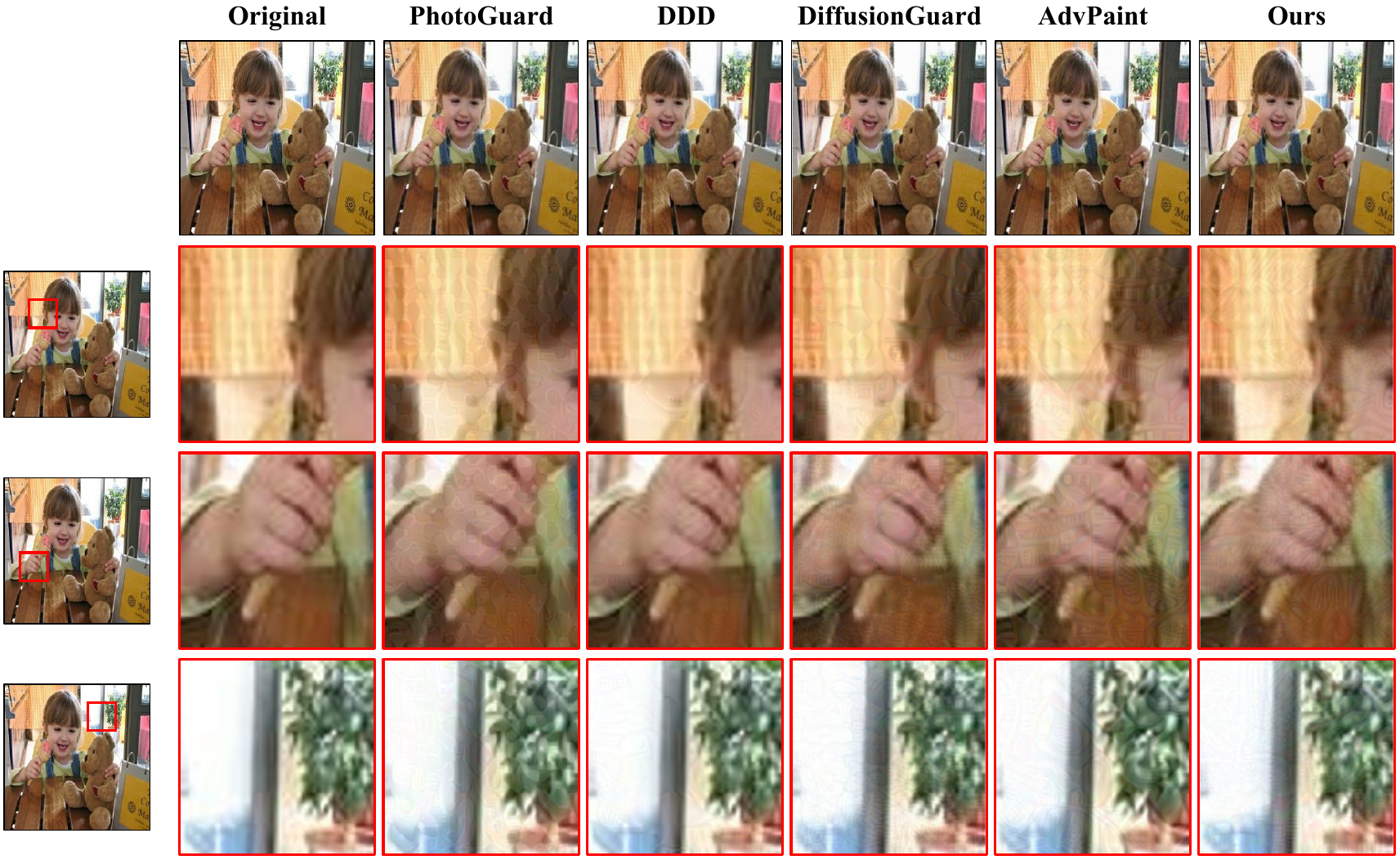}
  \caption{Qualitative evaluation of adversarial noise visibility. The red boxes in the first column indicate the regions that are magnified. Oracle, baselines, and our method are compared under the default setting of $\epsilon = 12/255$.}
  \Description{Noise appears in clean images and images protected by various methods.}
  \label{AFig: noise visualization}
\end{figure*}

To examine the visual characteristics of adversarial noise across different methods, we present a qualitative comparison in Fig. \ref{AFig: noise visualization}. This figure displays representative examples, with specific regions of each image magnified to facilitate closer inspection. The first column shows the original clean image which serves as a noise-free reference. The remaining columns present the outputs of each method under identical adversarial conditions.

All methods generate adversarial noise within a shared constraint of $\epsilon = 12/255$, ensuring a consistent and fair comparison. Under this noise budget, previous study~\cite{shan2023glaze} have shown that over 92\% of 1,156 surveyed artists indicated a willingness to publicly share protected images in place of the originals, underscoring the practical acceptability of this perturbation level. In our magnified views, the adversarial noise remains subtle and generally imperceptible to the naked eye. Although minor texture inconsistencies may occasionally emerge, these artifacts are minimal and do not compromise the overall perceptual realism of the images. This observation suggests that, under a controlled noise level, all methods maintain comparable visual quality without introducing conspicuous distortions.

\section{Extra Experiments}
\label{Sec: extra experiments}

\subsection{Attention Layer Selection}
\label{SubSec: Layer selection appendix}
\begin{table*}[ht]
    \centering
    \caption{Quantitative comparison from the ablation study on attention layer selection. The noise prediction model follows a U-Net architecture with four attention layers at spatial resolutions of 4096, 1024, 256, and 64. Excluding the largest-resolution layer ([1024, 256, 64]) yields the best alignment suppression. This suggests that the outermost layer primarily captures low-level visual details rather than prompt-relevant semantics. }
    \label{ATab: layer selection}
    \begin{tabular}{c c c c c c c c c c}
    \toprule
    \multirow{2}{*}{Selected layer sizes}
        & \multicolumn{2}{c}{CLIP Score $\downarrow$}
        & \multicolumn{2}{c}{Aesthetic $\downarrow$}
        & \multicolumn{2}{c}{PickScore $\downarrow$}
        & \multirow{2}{*}{LPIPS $\uparrow$}
        & \multirow{2}{*}{SSIM $\downarrow$}
        & \multirow{2}{*}{PSNR $\downarrow$} \\
    \cmidrule(lr){2-3}
    \cmidrule(lr){4-5}
    \cmidrule(lr){6-7}
    & \textit{all} & \textit{mask}
    & \textit{all} & \textit{mask}
    & \textit{all} & \textit{mask}
    &  &  &  \\

    \midrule
    $[4096, 1024, 256, 64]$ & 28.1688 & 22.4239 & \textbf{4.3683} & \textbf{4.0912} & 19.6864 & 18.7773 & \textbf{0.2583} & 0.6909 & 15.0347 \\
    $[4096, 1024, 256]$ & 28.2429 & 22.6715 & 4.3814 & 4.0991 & 19.6959 & 18.7799 & 0.2580 & 0.6915 & 14.9920 \\
    \rowcolor{backcolour}$[1024, 256, 64]$ \textbf{(\textit{ours})} & \textbf{27.3171} & \textbf{21.9557} & 4.4569 & 4.0916 & \textbf{19.5498} & \textbf{18.6491} & 0.2570 & \textbf{0.6894} & 14.9311 \\
    $[4096, 1024]$ & 30.3818 & 24.6188 & 4.4470 & 4.1522 & 20.1248 & 19.1412 & 0.2579 & 0.6936 & \textbf{14.6002} \\
    $[1024, 256]$ & 27.7441 & 22.2354 & 4.4673 & 4.0994 & 19.6100 & 18.6825 & 0.2578 & 0.6921 & 14.9843 \\
    $[256, 64]$ & 27.7084 & 22.0140 & 4.4720 & 4.1710 & 19.6136 & 18.7039 & 0.2567 & 0.6902 & 14.8592 \\
    \bottomrule
    \end{tabular}
\end{table*}

We conduct an ablation study to investigate the impact of attention layer selection on the effectiveness of PromptFlare. The noise prediction model in Stable Diffusion follows a U-Net architecture, comprising multiple attention layers operating at four distinct spatial resolutions: 4096, 1024, 256, and 64. Due to the symmetric nature of the U-Net, these layers are arranged in a downsampling and upsampling sequence, with the attention size decreasing from 4096 to 64 and then increasing back to 4096. To assess the contribution of each layer group, we construct the objective function in Eq. \ref{Eq: final loss} by selectively including different subsets of the attention layers and evaluate their impact on suppression performance.

As shown in Tab. \ref{ATab: layer selection}, two configurations exhibit superior performance: (1) using all attention layers and (2) excluding only the outermost layer. The former achieves the best results on image-level metrics such as Aesthetic Score and LPIPS, indicating stronger visual distortion but less prompt suppression. This suggests that the outermost layer captures overly fine-grained information that is irrelevant to the tokens of prompt, leading to image-level inconsistencies rather than alignment with the prompt. In contrast, excluding the outermost layer yields the best performance on semantic alignment metrics, including CLIP Score, PickScore, and SSIM. This configuration more effectively suppresses prompt influence without introducing unnecessary visual degradation. Given that our primary objective is to disrupt image-prompt alignment, we adopt the latter configuration as our default setting.

\subsection{Comparison on CFG Scale}
\label{Subsec: CFG scale appendix}

\begin{table*}[ht]
    \centering
    \caption{Quantitative evaluation of various methods on the EditBench dataset across different CFG scales. Evaluation metrics and all parameters except for the CFG scale follow the configurations specified in the main paper. The best results are highlighted in bold, and the second-best results are underlined. Our method consistently achieves the best performance in nearly all settings.}
    \label{ATab: cfg scale}
    \begin{tabular}{c | c c c c c c c c c c}
    \toprule
    CFG
        & \multirow{2}{*}{Methods}
        & \multicolumn{2}{c}{CLIP Score $\downarrow$}
        & \multicolumn{2}{c}{Aesthetic $\downarrow$}
        & \multicolumn{2}{c}{PickScore $\downarrow$}
        & \multirow{2}{*}{LPIPS $\uparrow$}
        & \multirow{2}{*}{SSIM $\downarrow$}
        & \multirow{2}{*}{PSNR $\downarrow$} \\
    \cmidrule(lr){3-4}
    \cmidrule(lr){5-6}
    \cmidrule(lr){7-8}
    scale & & \textit{all} & \textit{mask}
    & \textit{all} & \textit{mask}
    & \textit{all} & \textit{mask}
    &  &  &  \\
    \midrule
    \multirow{5}{*}{5.0} & Oracle
        & 31.7619 & 26.2522
        & 4.7458  & 4.3222
        & 20.6641 & 19.6067
        & --      & --      & --      \\
    & PhotoGuard~\cite{salman2023raising}
        & 31.6664 & 26.2743 & 4.5208 & 4.1972 & 20.2995 & 19.5250 & 0.2415 & 0.6905 & 15.2773 \\
    & DDD~\cite{son2024disrupting}
        & 30.3364 & 24.7067 & 4.5281 & 4.2377 & 20.1355 & 19.2843 & 0.2407 & 0.6952 & \textbf{14.8576} \\
    & DiffusionGuard~\cite{choi2024diffusionguard}
        & 30.1252 & 24.6623 & 4.5923 & 4.1537 & 20.2376 & 19.1576 & \underline{0.2541} & 0.6993 & 15.1116 \\
    & AdvPaint~\cite{jeon2025advpaint}
        & \underline{28.1244} & \underline{22.3805} & \textbf{4.2794} & \underline{4.1440} & \textbf{19.5301} & \underline{18.7200} & \textbf{0.2643} & \textbf{0.6824} & \underline{14.8830} \\
    & \cellcolor{backcolour}\SystemName~(\textit{\textbf{ours}})
        & \cellcolor{backcolour}\textbf{27.4556} & \cellcolor{backcolour}\textbf{21.9680} & \cellcolor{backcolour}\underline{4.4339} & \cellcolor{backcolour}\textbf{4.0835} & \cellcolor{backcolour}\underline{19.5619} & \cellcolor{backcolour}\textbf{18.6273} & \cellcolor{backcolour}0.2531 & \cellcolor{backcolour}\underline{0.6884} & \cellcolor{backcolour}15.2033 \\
    \midrule
    \multirow{5}{*}{7.5} & Oracle
        & 32.0412 & 26.6546
        & 4.7981  & 4.3493
        & 20.7639 & 19.6893
        & --      & --      & --      \\
    & PhotoGuard~\cite{salman2023raising}
        & 31.7589 & 26.3759
        & 4.5361  & 4.2780
        & 20.4062 & 19.6097
        & 0.2443  & 0.6947  & 14.8245 \\
    & DDD~\cite{son2024disrupting}
        & 30.9199 & 25.2935
        & 4.5625  & 4.5685
        & 20.2364 & 19.3621
        & 0.2449  & 0.6969  & \textbf{14.2616} \\
    & DiffusionGuard~\cite{choi2024diffusionguard}
        & 30.6709 & 25.1593
        & 4.6233  & 4.2097
        & 20.3921 & 19.3010
        & 0.2547  & 0.7016  & 14.5926 \\
    & AdvPaint~\cite{jeon2025advpaint}
        & \underline{28.4993} & \underline{22.8013} & \textbf{4.3205} & \underline{4.1715} & \underline{19.6327} & \underline{18.8247} & \textbf{0.2683} & \textbf{0.6839} & \underline{14.4043} \\
    & \cellcolor{backcolour}\SystemName~(\textit{\textbf{ours}})
        & \cellcolor{backcolour}\textbf{27.3171} & \cellcolor{backcolour}\textbf{21.9557}
        & \cellcolor{backcolour}\underline{4.4569}  & \cellcolor{backcolour}\textbf{4.0916}
        & \cellcolor{backcolour}\textbf{19.5498} & \cellcolor{backcolour}\textbf{18.6491}
        & \cellcolor{backcolour}\underline{0.2570}  & \cellcolor{backcolour}\underline{0.6894}  & \cellcolor{backcolour}14.9311 \\
        \midrule
    \multirow{5}{*}{10.0} & Oracle
        & 32.2078 & 26.9026 & 4.7714 & 4.3680 & 20.7482 & 19.6934
        & --      & --      & --      \\
    & PhotoGuard~\cite{salman2023raising}
        & 32.2236 & 26.8375 & 4.5845 & 4.3317 & 20.5524 & 19.6882 & 0.2436 & 0.6946 & 14.2426 \\
    & DDD~\cite{son2024disrupting}
        & 31.0983 & 25.5858 & 4.5496 & 4.2642 & 20.2251 & 19.4080 & 0.2460 & 0.6946 & \textbf{13.6427} \\
    & DiffusionGuard~\cite{choi2024diffusionguard}
        & 30.7826 & 25.2015 & 4.6081 & 4.2275 & 20.3860 & 19.3001 & 0.2553 & 0.6991 & 13.9538 \\
    & AdvPaint~\cite{jeon2025advpaint}
        & \underline{28.8520} & \underline{23.0349} & \textbf{4.3641} & \underline{4.1996} & \underline{19.6356} & \underline{18.8641} & \textbf{0.2693} & \textbf{0.6827} & \underline{13.7812} \\
    & \cellcolor{backcolour}\SystemName~(\textit{\textbf{ours}})
        & \cellcolor{backcolour}\textbf{27.6102} & \cellcolor{backcolour}\textbf{22.2388} & \cellcolor{backcolour}\underline{4.4657} & \cellcolor{backcolour}\textbf{4.1380} & \cellcolor{backcolour}\textbf{19.6228} & \cellcolor{backcolour}\textbf{18.6797} & \cellcolor{backcolour}\underline{0.2574} & \cellcolor{backcolour}\underline{0.6889} & \cellcolor{backcolour}14.4737 \\
        \midrule
    \multirow{5}{*}{12.5} & Oracle
        & 32.5193 & 27.1049 & 4.8065 & 4.3630 & 20.8273 & 19.7422
        & --      & --      & --      \\
    & PhotoGuard~\cite{salman2023raising}
        & 32.2687 & 27.0028 & 4.6076 & 4.3515 & 20.6296 & 19.7636 & 0.2451 & 0.6960 & 14.0523 \\
    & DDD~\cite{son2024disrupting}
        & 31.2946 & 25.9218 & 4.5837 & 4.2848 & 20.3358 & 19.4966 & 0.2465 & 0.6978 & \underline{13.5080} \\
    & DiffusionGuard~\cite{choi2024diffusionguard}
        & 31.2414 & 25.5613 & 4.6782 & 4.2866 & 20.4420 & 19.3824 & 0.2564 & 0.7012 & 13.7455 \\
    & AdvPaint~\cite{jeon2025advpaint}
        & \underline{28.9791} & \underline{23.3433} & \textbf{4.4246} & \underline{4.2837} & \underline{19.6990} & \underline{18.9127} & \textbf{0.2713} & \textbf{0.6826} & \textbf{13.4253} \\
    & \cellcolor{backcolour}\SystemName~(\textit{\textbf{ours}})
        & \cellcolor{backcolour}\textbf{27.8737} & \cellcolor{backcolour}\textbf{22.3322} & \cellcolor{backcolour}\underline{4.4670} & \cellcolor{backcolour}\textbf{4.1183} & \cellcolor{backcolour}\textbf{19.6833} & \cellcolor{backcolour}\textbf{18.7293} & \cellcolor{backcolour}\underline{0.2610} & \cellcolor{backcolour}\underline{0.6885} & \cellcolor{backcolour}14.2170 \\
        \midrule
    \multirow{5}{*}{15.0} & Oracle
        & 32.5134 & 27.1428 & 4.7955 & 4.3922 & 20.8454 & 19.7851
        & --      & --      & --      \\
    & PhotoGuard~\cite{salman2023raising}
        & 32.5526 & 27.3475 & 4.6245 & 4.4023 & 20.5984 & 19.7948 & 0.2434 & 0.6964 & 13.7564 \\
    & DDD~\cite{son2024disrupting}
        & 31.6285 & 26.1495 & 4.5961 & 4.3251 & 20.3729 & 19.5076 & 0.2453 & 0.6978 & \underline{13.1733} \\
    & DiffusionGuard~\cite{choi2024diffusionguard}
        & 31.4514 & 25.7216 & 4.6420 & 4.2756 & 20.4914 & 19.4286 & 0.2568 & 0.6985 & 13.4346 \\
    & AdvPaint~\cite{jeon2025advpaint}
        & \underline{29.3454} & \underline{23.6542} & \textbf{4.3996} & \underline{4.2679} & \underline{19.7370} & \underline{18.9771} & \textbf{0.2711} & \textbf{0.6809} & \textbf{13.0538} \\
    & \cellcolor{backcolour}\SystemName~(\textit{\textbf{ours}})
        & \cellcolor{backcolour}\textbf{28.0471} & \cellcolor{backcolour}\textbf{22.4716} & \cellcolor{backcolour}\underline{4.4617} & \cellcolor{backcolour}\textbf{4.1173} & \cellcolor{backcolour}\textbf{19.6656} & \cellcolor{backcolour}\textbf{18.7445} & \cellcolor{backcolour}\underline{0.2607} & \cellcolor{backcolour}\underline{0.6876} & \cellcolor{backcolour}13.9681 \\
    \bottomrule
    \end{tabular}
\end{table*}

In this section, we present a comprehensive evaluation of our method on the EditBench dataset under varying CFG scales, which directly affect the diffusion generation process. Tab. \ref{ATab: cfg scale} reports the quantitative results as the CFG scale is progressively increased from 5.0 to 15.0. To ensure fair comparison, all experimental settings except for the CFG scale adhere to the configurations specified in the main paper.

Our method consistently outperforms all baselines across nearly all CFG scales. In particular, it achieves substantial improvements on reference-free metrics, often exceeding the performance gap observed between baselines and the Oracle. This performance advantage persists even as the CFG scale increases. Although our method yields significantly lower reference-free metric scores compared to other approaches, it also demonstrates the smallest increase in these scores as the CFG scale intensifies, underscoring its robustness against stronger generative signals. Regarding reference-based image quality metrics, our method maintains second-best performance in most cases.

Overall, the results indicate that our method effectively disrupts the conditional sampling process, in alignment with the objective function specifically designed to suppress prompt adherence.

\subsection{Visualize Attention Map}
\label{Subsec: attention map appendix}

\begin{figure*}[th]
  \centering
  \includegraphics[width=\linewidth]{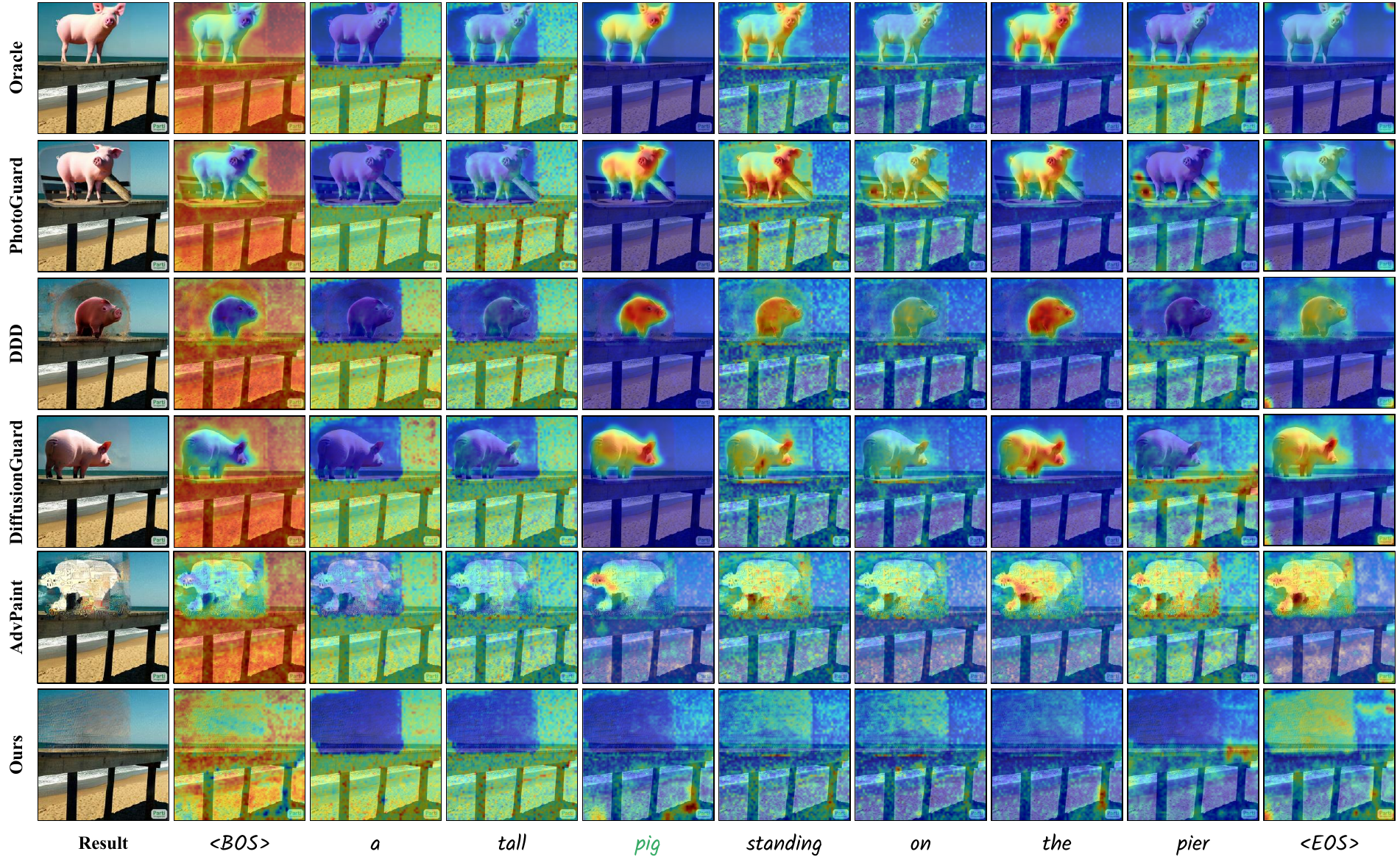}
  \caption{Token-wise attention maps of inpainting results generated across various methods. The green token indicates the core object, while <BOS> and <EOS> represent the beginning-of-sequence (BOS) token and end-of-sequence (EOS) token, respectively. While existing baselines exhibit attention map similar to the Oracle, our method successfully suppresses the contribution of tokens except BOS token.}
  \label{AFig: attention map}
  \Description{attention maps activation figure.}
\end{figure*}

In this section, we extend the analysis presented in Sec. \ref{Subsec: attention map} by applying the same attention visualization procedure to baseline methods. To better understand how different approaches manage prompt influence during inpainting, we employ DAAM~\cite{tang2023daam} to generate token-wise attention maps that highlight which parts of the prompt guide the generation process and where each token exerts influence within the image.

As shown in Fig. \ref{AFig: attention map}, the Oracle model which represents the standard, unaltered diffusion process exhibits a strong correlation between the inpainted region and the prompt token corresponding to the core object, highlighted in green. This indicates that the core object token plays a dominant role in guiding generation, as expected in conventional conditional sampling settings.

Interestingly, baseline methods display attention patterns that closely resemble those of the Oracle. Despite adopting different loss functions aimed at modulating inpainting behavior, these methods fail to substantially diminish the prompt’s influence. This is primarily because they penalize discrepancies only at the image level and do not explicitly suppress the model’s attention to the core object token. As a result, the inpainting outputs remain tightly aligned with the semantics of the original prompt, showing little deviation from Oracle.

Among the baseline methods, AdvPaint is the only approach that visually deviates from the Oracle, appearing to generate the generation of unnatural white artifacts and distorted color patterns. However, token-wise attention maps reveal that it exhibit similar trends to other baselines, with strong activations concentrated on the core object token and the EOS token. This indicates that, despite its visual artifacts, AdvPaint does not effectively disrupt the prompt-image alignment. Therefore, increasing the CFG scale to amplify conditional sampling would likely bypass the defense and result in inpainted outputs that closely follow the intended prompt.

In contrast, our method is explicitly designed to suppress prompt influence during the conditional sampling process. As demonstrated in earlier experiments and further corroborated by this visualization, our approach uniquely disrupts the attention dynamics such that prompt tokens—including those representing the core object—exert minimal to no influence on the inpainted region. This qualitative evidence reinforces the efficacy of our method in reducing prompt fidelity.

\subsection{Unseen Masks}

\begin{figure*}[th]
  \centering
  \includegraphics[width=\linewidth]{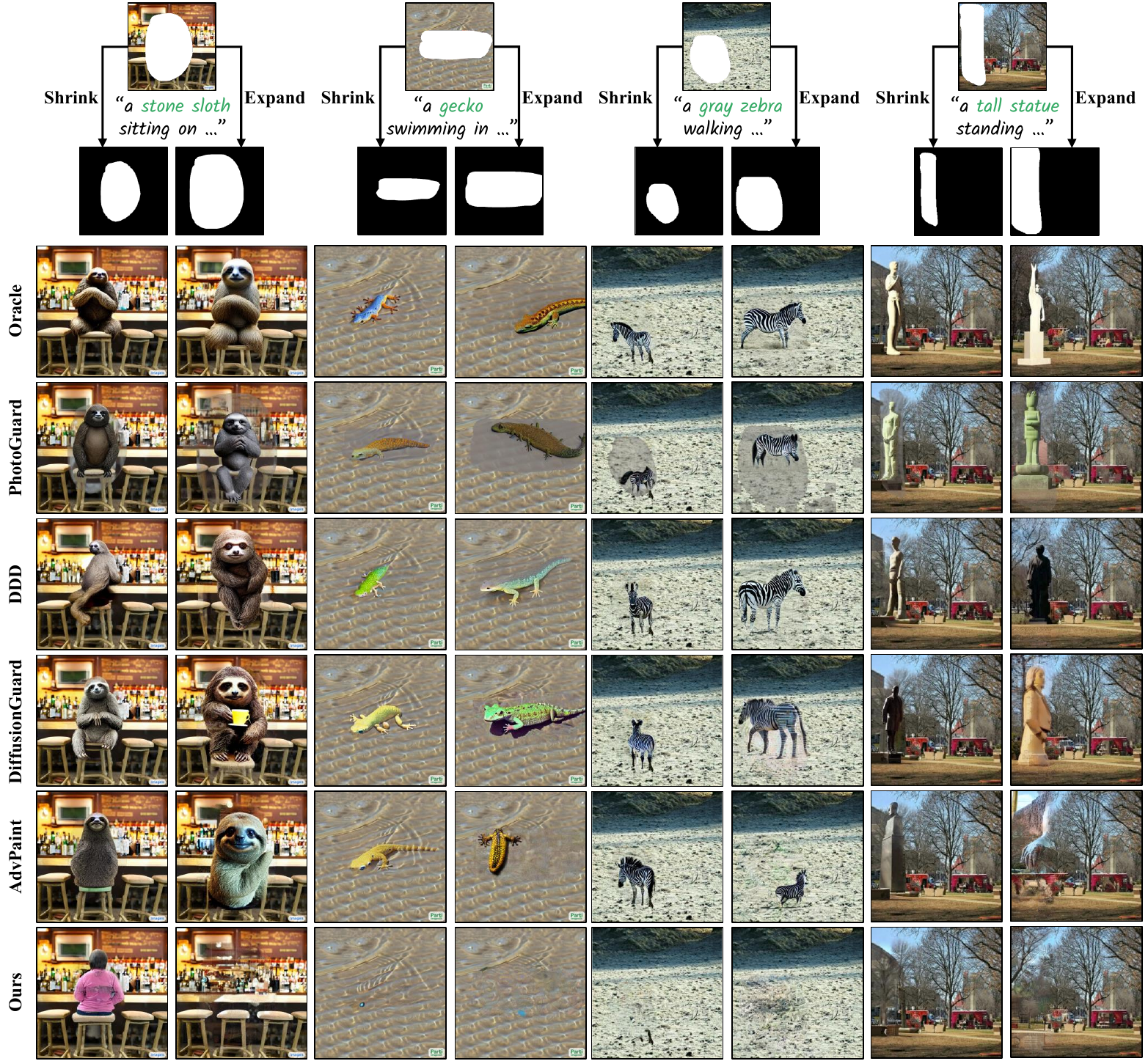}
  \caption{Qualitative evaluation of various methods on the EditBench dataset under both shrunk and expanded masks. Our method demonstrates strong generalization by effectively suppressing generation even when presented with previously unseen mask variations during inference.}
  \Description{A woman and a girl in white dresses sit in an open car.}
  \label{AFig: unseen mask}
\end{figure*}

\begin{table*}[th]
    \centering
    \caption{Quantitative evaluation of various methods on the EditBench dataset under both shrunk and expanded masks. Evaluation metrics and all parameters follow the configurations specified in the main paper. The best results are highlighted in bold, and the second-best results are underlined. Our method consistently achieves the best performance in nearly all settings.}
    \label{ATab: unseen mask}
    \begin{tabular}{c | c c c c c c c c c c}
    \toprule
    Unseen
        & \multirow{2}{*}{Methods}
        & \multicolumn{2}{c}{CLIP Score $\downarrow$}
        & \multicolumn{2}{c}{Aesthetic $\downarrow$}
        & \multicolumn{2}{c}{PickScore $\downarrow$}
        & \multirow{2}{*}{LPIPS $\uparrow$}
        & \multirow{2}{*}{SSIM $\downarrow$}
        & \multirow{2}{*}{PSNR $\downarrow$} \\
    \cmidrule(lr){3-4}
    \cmidrule(lr){5-6}
    \cmidrule(lr){7-8}
    mask & & \textit{all} & \textit{mask}
    & \textit{all} & \textit{mask}
    & \textit{all} & \textit{mask}
    &  &  &  \\
    \midrule
    \multirow{5}{*}{shrink} & Oracle
        & 32.5933 & 26.6302 & 4.9513 & 4.3910 & 20.9632 & 19.7350
        & --      & --      & --      \\
    & PhotoGuard~\cite{salman2023raising}
        & 32.4123 & 26.5792 & \textbf{4.5050} & 4.2146 & 20.6759 & 19.7361 & 0.3275 & \underline{0.6082} & 14.0819 \\
    & DDD~\cite{son2024disrupting}
        & 31.8734 & 25.9618 & 4.7568 & 4.3384 & 20.7451 & 19.5670 & 0.3163 & 0.6207 & \underline{13.3597} \\
    & DiffusionGuard~\cite{choi2024diffusionguard}
        & \underline{30.8146} & \underline{25.1447} & 4.6219 & \underline{4.2092} & \underline{20.2961} & \underline{19.2730} & \underline{0.3369} & 0.6147 & 13.6111 \\
    & AdvPaint~\cite{jeon2025advpaint}
        & 31.4950 & 25.6807 & \underline{4.5706} & 4.4083 & 20.4393 & 19.5026 & \textbf{0.3449} & \textbf{0.6022} & \textbf{12.9658} \\
    & \cellcolor{backcolour}\SystemName~(\textit{\textbf{ours}})
        & \cellcolor{backcolour}\textbf{29.7535} & \cellcolor{backcolour}\textbf{24.1677} & \cellcolor{backcolour}4.5998 & \cellcolor{backcolour}\textbf{4.1937} & \cellcolor{backcolour}\textbf{20.1276} & \cellcolor{backcolour}\textbf{19.1445} & \cellcolor{backcolour}0.3280 & \cellcolor{backcolour}0.6141 & \cellcolor{backcolour}13.6011 \\
    \midrule
    \multirow{5}{*}{expand} & Oracle
        & 32.1247 & 27.0379 & 4.8460 & 4.3189 & 20.9724 & 19.7724
        & --      & --      & --      \\
    & PhotoGuard~\cite{salman2023raising}
        & 32.2247 & 26.8645 & \textbf{4.6107} & \textbf{4.2196} & \underline{20.7289} & 19.6617 & \textbf{0.1730} & \textbf{0.7877} & 16.5412 \\
    & DDD~\cite{son2024disrupting}
        & 32.3156 & 26.7593 & 4.7975 & 4.3036 & 20.9069 & 19.7389 & 0.1642 & 0.7966 & 16.0091 \\
    & DiffusionGuard~\cite{choi2024diffusionguard}
        & \underline{31.9134} & \underline{26.4790} & 4.7895 & 4.3269 & 20.8746 & 19.6784 & 0.1653 & 0.7986 & 16.0813 \\
    & AdvPaint~\cite{jeon2025advpaint}
        & 31.9592 & 25.5321 & 4.7853 & 4.3798 & 20.7690 & \underline{19.6573} & \underline{0.1708} & \underline{0.7918} & \textbf{15.6843} \\
    & \cellcolor{backcolour}\SystemName~(\textit{\textbf{ours}})
        & \cellcolor{backcolour}\textbf{30.8093} & \cellcolor{backcolour}\textbf{25.2906} & \cellcolor{backcolour}\underline{4.7717} & \cellcolor{backcolour}\underline{4.2802} & \cellcolor{backcolour}\textbf{20.5666} & \cellcolor{backcolour}\textbf{19.3913} & \cellcolor{backcolour}0.1657 & \cellcolor{backcolour}0.7948 & \cellcolor{backcolour}\underline{15.9663} \\
    \bottomrule
    \end{tabular}
\end{table*}

In this section, we evaluate the degree to which different methods overfit to the input mask by evaluating their performance when the original adversarial mask is altered. Specifically, we test each method under conditions where the mask used during the adversarial noise generation is either shrunk or expanded before being fed as input for inpainting. This evaluation is critical, as it is unrealistic to assume that malicious users will always supply the same mask. Therefore, generalization to unseen mask variations is essential for practical robustness.

To simulate such variations, we apply morphological transformations to the original mask using erosion and dilation with a kernel size of 5 and 10 iterations.

Fig. \ref{AFig: unseen mask} illustrated qualitative comparisons of inpainting results under these unseen masks. Existing methods fail to maintain their suppression capabilities when the mask is modified, indicating a tendency to overfit to the original mask used during adversarial attack. Notably, DiffusionGuard~\cite{choi2024diffusionguard}, despite incorporating random boundary shrinking during its noise optimization process, also struggles to generalize to these new mask conditions.

Tab. \ref{ATab: unseen mask} provides a quantitative comparison under the same evaluation protocol. Our method consistently achieves either the best or second-best performance across most metrics, regardless of whether the mask is shrunk or expanded. In particular, for CLIP Score and PickScore—metrics that reflect the semantic alignment between the generated image and the prompt—our method achieves the best performance in all cases. These results clearly indicate that our approach does not overfit to a specific mask but instead generalizes effectively across varying inpainting regions.

\subsection{Model Transferability}

\begin{figure*}[th]
  \centering
  \includegraphics[width=0.85\linewidth]{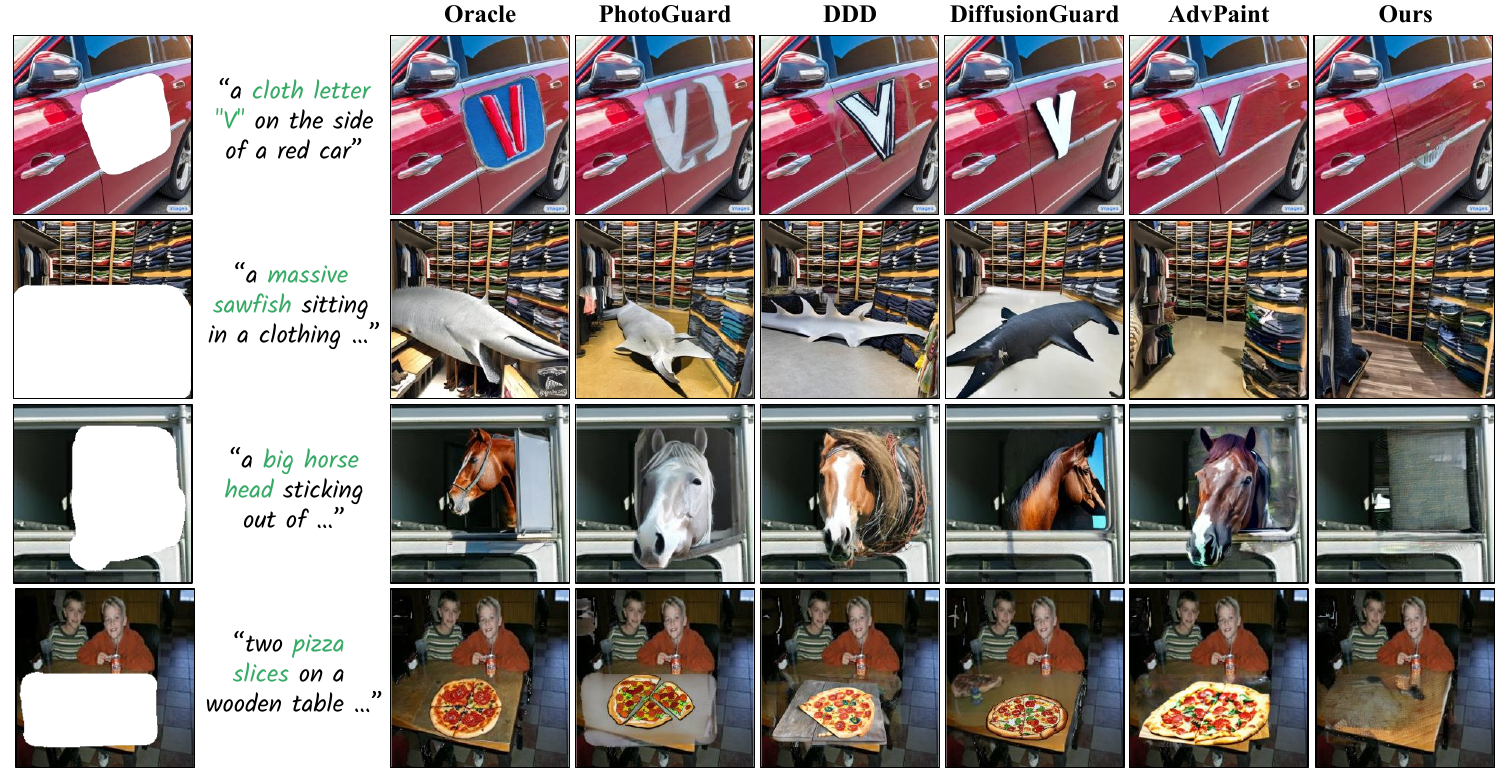}
  \caption{Qualitative evaluation of various methods on the EditBench dataset under using StabilityAI's Stable Diffusion 2 Inapinting model. Compared to other baselines, our method exhibits notably stronger transferability across models.}
  \Description{A woman and a girl in white dresses sit in an open car.}
  \label{AFig: model transferability}
\end{figure*}

\begin{table*}[th]
    \centering
    \caption{Quantitative evaluation of various methods on the EditBench dataset under using StabilityAI's Stable Diffusion 2 Inapinting model. Evaluation metrics and all parameters follow the configurations specified in the main paper. The best results are highlighted in bold, and the second-best results are underlined. Our method consistently achieves the best performance in nearly all settings.}
    \label{ATab: model transferability}
    \begin{tabular}{c c c c c c c c c c}
    \toprule
    \multirow{2}{*}{Methods}
        & \multicolumn{2}{c}{CLIP Score $\downarrow$}
        & \multicolumn{2}{c}{Aesthetic $\downarrow$}
        & \multicolumn{2}{c}{PickScore $\downarrow$}
        & \multirow{2}{*}{LPIPS $\uparrow$}
        & \multirow{2}{*}{SSIM $\downarrow$}
        & \multirow{2}{*}{PSNR $\downarrow$} \\
    \cmidrule(lr){2-3}
    \cmidrule(lr){4-5}
    \cmidrule(lr){6-7}
    & \textit{all} & \textit{mask}
    & \textit{all} & \textit{mask}
    & \textit{all} & \textit{mask}
    &  &  &  \\
    \midrule
    Oracle
        & 32.5839 & 27.4403 & 4.7508 & 4.1858 & 20.9585 & 19.7839
        & --      & --      & --      \\
    PhotoGuard~\cite{salman2023raising}
        & 32.8590 & 27.6234 & 4.5641 & 4.2867 & 20.7625 & 19.8975 & 0.2431 & \underline{0.6935} & 14.6744 \\
    DDD~\cite{son2024disrupting}
        & 32.4397 & 26.8188 & 4.6872 & 4.2946 & 20.8405 & 19.7605 & 0.2392 & 0.6991 & \underline{14.1042} \\
    DiffusionGuard~\cite{choi2024diffusionguard}
        & 31.8598 & 26.2592 & 4.6644 & 4.2257 & 20.7498 & 19.6185 & 0.2506 & 0.7003 & 14.3195 \\
    AdvPaint~\cite{jeon2025advpaint}
        & \underline{31.1940} & \underline{25.2554} & \textbf{4.4852} & \underline{4.3121} & \textbf{20.3614} & \underline{19.4088} & \textbf{0.2622} & \textbf{0.6878} & \textbf{14.0053} \\
    \rowcolor{backcolour}\SystemName~(\textit{\textbf{ours}})
        & \textbf{31.1550} & \textbf{25.2054} & \underline{4.5610} & \textbf{4.2194} & \underline{20.4748} & \textbf{19.3669} & \underline{0.2531} & \underline{0.6935} & 14.3442 \\
    \bottomrule
    \end{tabular}
\end{table*}

In this section, we evaluate the transferability of adversarial attack methods originally developed using RunwayML’s Stable Diffusion Inpainting model, by applying them to a different inpainting architecture. As discussed in Sec. \ref{Subsec: Problem statement}, we assume that a malevolent user may exploit publicly available inpainting models for malcious manipulation. Given this assumption, it is also plausible that such users might utilize alternative models. To this end, we evaluate the performance of each method using StabilityAI’s Stable Diffusion 2 Inpainting model~\cite{rombach2022high}.

Fig. \ref{AFig: model transferability} illustrates a qualitative comparison under this new setting. Although the same prompts and masks used in previous experiments are employed, existing methods fail to suppress content generation once the model is changed. In some instances, gray blurring artifacts appear, which are clear indicators of attempted generation suppression based on image-level inconsistency, yet the core object remains sharply overlaid. This provides direct evidence that these methods fail to inhibit content generation despite apparent suppression efforts. In contrast, our method consistently inhibits content generation, demonstrating robustness even under model variation.

Tab. \ref{ATab: model transferability} provides the corresponding quantitative results. While the margin of improvement in CLIP-based metrics is less pronounced compared to earlier experiments, our method continues to achieve the best performance across most metrics, further validating its effectiveness and generalization capability across different inpainting models.

\subsection{Comparison on Noise Budget}
\label{Subsec: noise budget}

\begin{figure*}[th]
  \centering
  \includegraphics[width=0.75\linewidth]{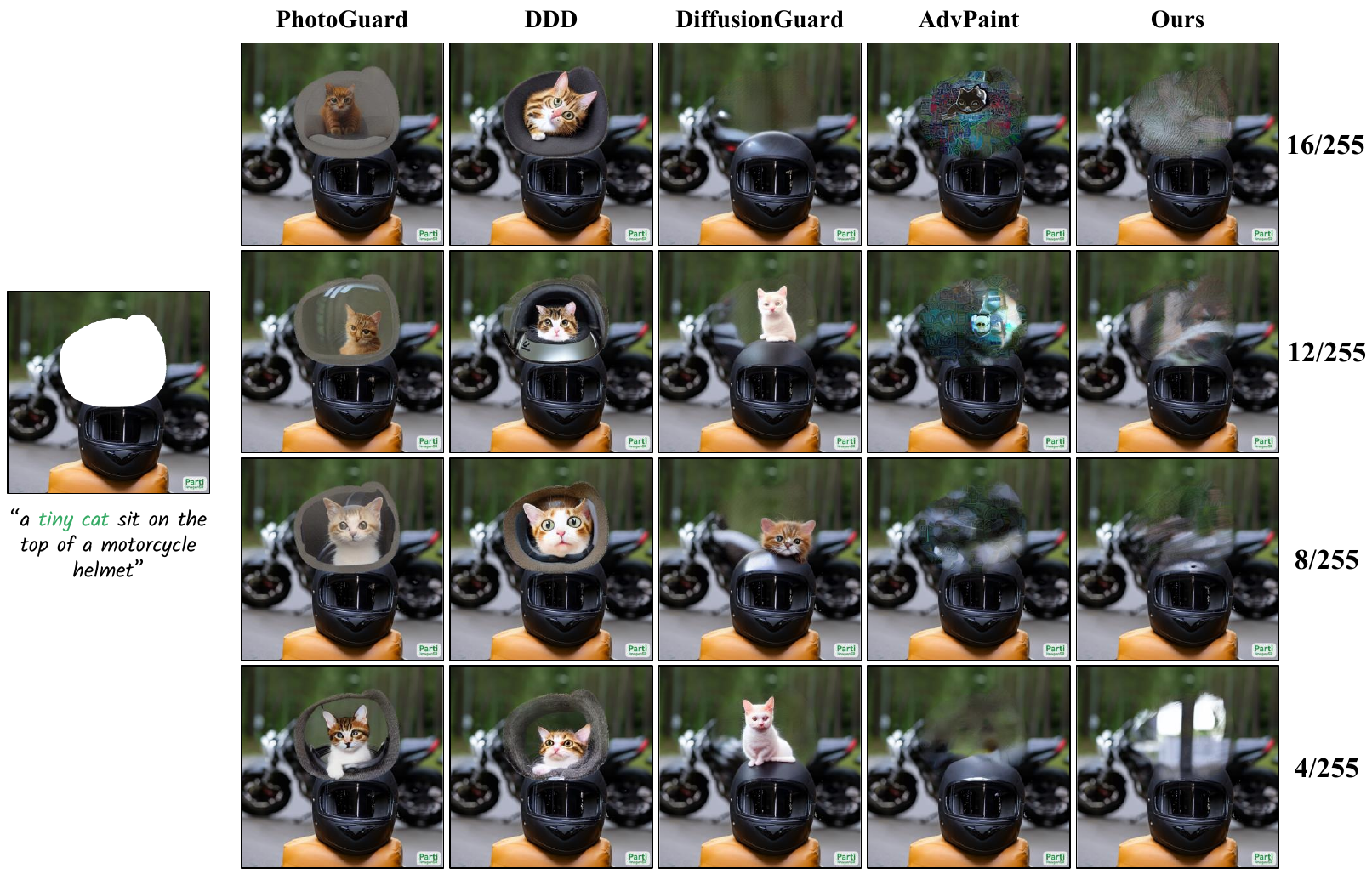}
  \caption{Qualitative evaluation of various methods on the EditBench dataset across different noise budgets. Our method consistently suppresses content generation even at low noise levels, demonstrating strong robustness to noise budget constraints.}
  \Description{A woman and a girl in white dresses sit in an open car.}
  \label{AFig: noise budget}
\end{figure*}

\begin{table*}[th]
    \centering
    \caption{Quantitative evaluation of various methods on the EditBench dataset across different noise budgets. Evaluation metrics and all parameters except for the noise budget follow the configurations specified in the main paper. The best results are highlighted in bold, and the second-best results are underlined. Our method consistently achieves the best performance in nearly all settings.}
    \label{ATab: noise budget}
    \begin{tabular}{c | c c c c c c c c c c}
    \toprule
    noise
        & \multirow{2}{*}{Methods}
        & \multicolumn{2}{c}{CLIP Score $\downarrow$}
        & \multicolumn{2}{c}{Aesthetic $\downarrow$}
        & \multicolumn{2}{c}{PickScore $\downarrow$}
        & \multirow{2}{*}{LPIPS $\uparrow$}
        & \multirow{2}{*}{SSIM $\downarrow$}
        & \multirow{2}{*}{PSNR $\downarrow$} \\
    \cmidrule(lr){3-4}
    \cmidrule(lr){5-6}
    \cmidrule(lr){7-8}
    budget & & \textit{all} & \textit{mask}
    & \textit{all} & \textit{mask}
    & \textit{all} & \textit{mask}
    &  &  &  \\
    \midrule
    \multirow{5}{*}{16/255} & Oracle
        & 32.0412 & 26.6546 & 4.7981 & 4.3493 & 20.7639 & 19.6893
        & --      & --      & --      \\
    & PhotoGuard~\cite{salman2023raising}
        & 31.8315 & 26.4262 & 4.4865 & 4.2136 & 20.4002 & 19.5879 & 0.2452 & 0.6944 & 15.0890 \\
    & DDD~\cite{son2024disrupting}
        & 30.5712 & 24.8876 & 4.4942 & 4.2389 & 20.1674 & 19.2784 & 0.2468 & 0.6937 & \underline{14.3690} \\
    & DiffusionGuard~\cite{choi2024diffusionguard}
        & 30.7015 & 25.1550 & 4.5940 & 4.1766 & 20.3594 & 19.2718 & 0.2562 & 0.6997 & 14.5944 \\
    & AdvPaint~\cite{jeon2025advpaint}
        & \underline{28.2006} & \underline{22.4697} & \textbf{4.2765} & \underline{4.1440} & \textbf{19.4995} & \underline{18.7295} & \textbf{0.2728} & \textbf{0.6802} & \textbf{14.2763} \\
    & \cellcolor{backcolour}\SystemName~(\textit{\textbf{ours}})
        & \cellcolor{backcolour}\textbf{27.2861} & \cellcolor{backcolour}\textbf{21.9937} & \cellcolor{backcolour}\underline{4.4217} & \cellcolor{backcolour}\textbf{4.0903} & \cellcolor{backcolour}\underline{19.5458} & \cellcolor{backcolour}\textbf{18.6345} & \cellcolor{backcolour}\underline{0.2593} & \cellcolor{backcolour}\underline{0.6862} & \cellcolor{backcolour}15.0041 \\
    \midrule
    \multirow{5}{*}{12/255} & Oracle
        & 32.0412 & 26.6546
        & 4.7981  & 4.3493
        & 20.7639 & 19.6893
        & --      & --      & --      \\
    & PhotoGuard~\cite{salman2023raising}
        & 31.7589 & 26.3759
        & 4.5361  & 4.2780
        & 20.4062 & 19.6097
        & 0.2443  & 0.6947  & 14.8245 \\
    & DDD~\cite{son2024disrupting}
        & 30.9199 & 25.2935
        & 4.5625  & 4.5685
        & 20.2364 & 19.3621
        & 0.2449  & 0.6969  & \textbf{14.2616} \\
    & DiffusionGuard~\cite{choi2024diffusionguard}
        & 30.6709 & 25.1593
        & 4.6233  & 4.2097
        & 20.3921 & 19.3010
        & 0.2547  & 0.7016  & 14.5926 \\
    & AdvPaint~\cite{jeon2025advpaint}
        & \underline{28.4993} & \underline{22.8013} & \textbf{4.3205} & \underline{4.1715} & \underline{19.6327} & \underline{18.8247} & \textbf{0.2683} & \textbf{0.6839} & \underline{14.4043} \\
    & \cellcolor{backcolour}\SystemName~(\textit{\textbf{ours}})
        & \cellcolor{backcolour}\textbf{27.3171} & \cellcolor{backcolour}\textbf{21.9557}
        & \cellcolor{backcolour}\underline{4.4569}  & \cellcolor{backcolour}\textbf{4.0916}
        & \cellcolor{backcolour}\textbf{19.5498} & \cellcolor{backcolour}\textbf{18.6491}
        & \cellcolor{backcolour}\underline{0.2570}  & \cellcolor{backcolour}\underline{0.6894}  & \cellcolor{backcolour}14.9311 \\
        \midrule
    \multirow{5}{*}{8/255} & Oracle
        & 32.0412 & 26.6546 & 4.7981 & 4.3493 & 20.7639 & 19.6893
        & --      & --      & --      \\
    & PhotoGuard~\cite{salman2023raising}
        & 32.2413 & 27.0170 & 4.6124 & 4.2852 & 20.5743 & 19.6755 & 0.2382 & 0.6976 & 14.6330 \\
    & DDD~\cite{son2024disrupting}
        & 31.2583 & 25.5705 & 4.5715 & 4.2755 & 20.3193 & 19.4164 & 0.2426 & 0.6981 & \textbf{14.2483} \\
    & DiffusionGuard~\cite{choi2024diffusionguard}
        & 30.4893 & 24.9279 & 4.7010 & 4.2413 & 20.4564 & 19.3274 & 0.2528 & 0.7035 & 14.6271 \\
    & AdvPaint~\cite{jeon2025advpaint}
        & \underline{28.9715} & \underline{23.4405} & \textbf{4.3588} & \underline{4.1744} & \underline{19.7618} & \underline{18.9179} & \textbf{0.2625} & \textbf{0.6883} & \underline{14.3249} \\
    & \cellcolor{backcolour}\SystemName~(\textit{\textbf{ours}})
        & \cellcolor{backcolour}\textbf{27.7435} & \cellcolor{backcolour}\textbf{22.2758} & \cellcolor{backcolour}\underline{4.4863} & \cellcolor{backcolour}\textbf{4.0689} & \cellcolor{backcolour}\textbf{19.6667} & \cellcolor{backcolour}\textbf{18.7310} & \cellcolor{backcolour}\underline{0.2544} & \cellcolor{backcolour}\underline{0.6942} & \cellcolor{backcolour}14.8508 \\
        \midrule
    \multirow{5}{*}{4/255} & Oracle
        & 32.0412 & 26.6546 & 4.7981 & 4.3493 & 20.7639 & 19.6893
        & --      & --      & --      \\
    & PhotoGuard~\cite{salman2023raising}
        & 32.2201 & 26.9377 & 4.6867 & 4.3038 & 20.6820 & 19.7053 & 0.2327 & 0.7031 & 14.5183 \\
    & DDD~\cite{son2024disrupting}
        & 31.6162 & 26.0290 & \underline{4.6712} & 4.3212 & 20.4433 & 19.5250 & 0.2361 & 0.7035 & \textbf{14.1426} \\
    & DiffusionGuard~\cite{choi2024diffusionguard}
        & 30.8760 & 25.3622 & 4.7937 & 4.2734 & 20.5815 & 19.4050 & 0.2504 & 0.7045 & 14.4446 \\
    & AdvPaint~\cite{jeon2025advpaint}
        & \underline{30.3161} & \underline{24.4657} & \textbf{4.6147} & \underline{4.2636} & \underline{20.2124} & \underline{19.1945} & \textbf{0.2561} & \textbf{0.6974} & \underline{14.2692} \\
    & \cellcolor{backcolour}\SystemName~(\textit{\textbf{ours}})
        & \cellcolor{backcolour}\textbf{28.9768} & \cellcolor{backcolour}\textbf{23.4022} & \cellcolor{backcolour}4.6830 & \cellcolor{backcolour}\textbf{4.1961} & \cellcolor{backcolour}\textbf{20.0529} & \cellcolor{backcolour}\textbf{18.9809} & \cellcolor{backcolour}\underline{0.2520} & \cellcolor{backcolour}\underline{0.7005} & \cellcolor{backcolour}14.5139 \\
    \bottomrule
    \end{tabular}
\end{table*}

In this section, we evaluate the effectiveness of adversarial noise injection across varying noise budgets, with $\epsilon$ ranging from 4/255 to 16/255. These values are adopted from standard configurations used in previous study.

Fig. \ref{AFig: noise budget} illustrates a qualitative comparison of inpainting results under different noise budgets. Since the noise budget only pertains to adversarial attack, the Oracle model remains unchanged and is therefore excluded from the comparison. Notably, our method successfully suppresses content generation even at the smallest noise level, whereas the performance of the baseline methods degrades rapidly as the noise budget decreases. This indicates that our objective function is more precisely aligned with the goal of generation suppression, allowing the method to remain effective even when the influence of adversarial noise is reduced.

Tab. \ref{ATab: noise budget} provides a quantitative analysis across the same range of noise budgets. Consistent with previous findings, our method achieves a substantial lead in CLIP Score across all noise budgets and outperforms other approaches on most evaluation metrics, demonstrating its robustness even under strict noise constraints.

\subsection{Comparison on Inference Step}

In this section, we evaluate the performance of each method across varying the number of inference steps from 100 to 50 during inpainting. In adversarial attacks for generation suppression, it is common to set the number of inference steps to 4 during adversarial noise computation and extract gradients from part or all of the diffusion trajectory. This introduces the risk of overfitting to a specific inference configuration.

\begin{figure*}[th]
  \centering
  \includegraphics[width=\linewidth]{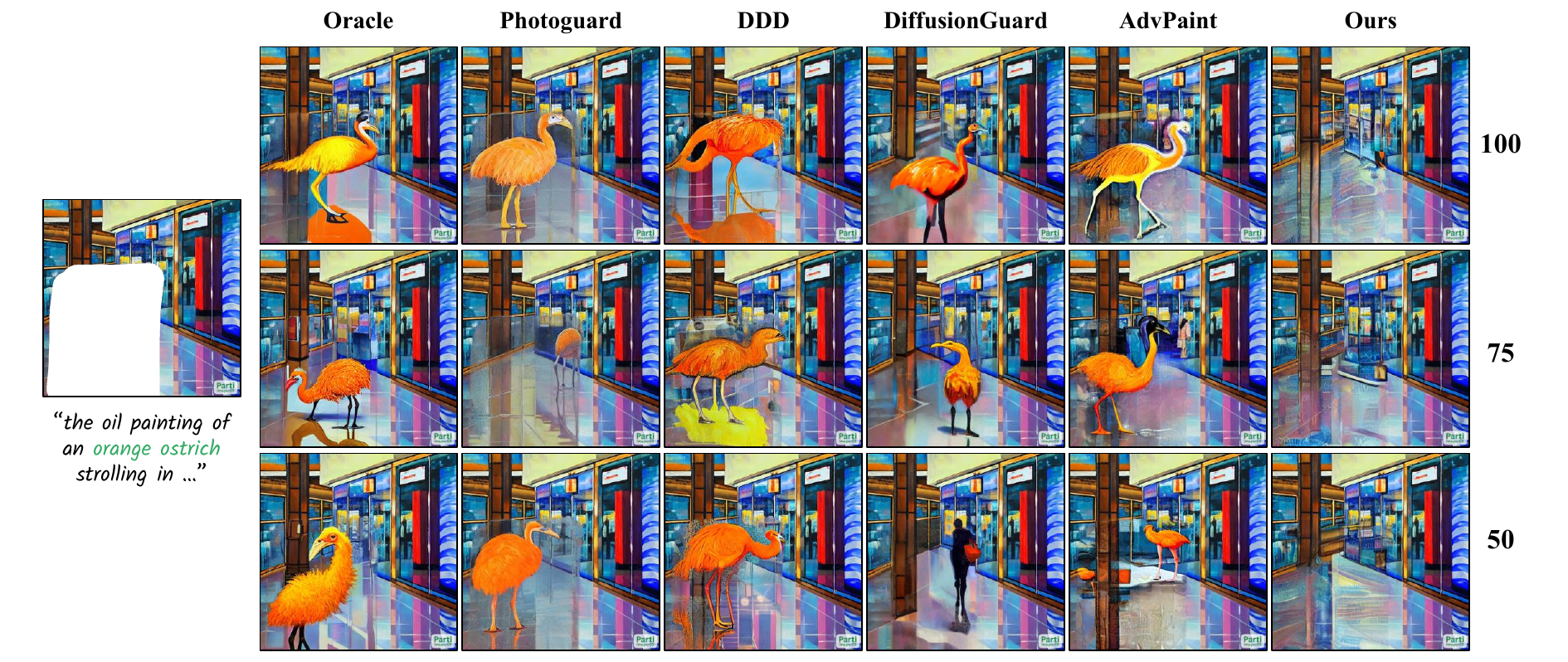}
  \caption{Qualitative evaluation of various methods on the EditBench dataset across different inference steps. Unlike the baselines, our method consistently achieves effective generation suppression across all inference step settings.}
  \Description{A woman and a girl in white dresses sit in an open car.}
  \label{AFig: inference step}
\end{figure*}

Fig. \ref{AFig: inference step} illustrates a qualitative comparison across different inference steps. In contrast to baseline methods, our approach consistently suppresses content generation across all inference step settings. However, it remains unclear whether this is due to robustness against variation in the inference step or is simply attributable to the overall superiority of our method.

The results in the Tab. \ref{ATab: inference step} further support this analysis. When comparing quantitative performance across varying inference steps, we observe minimal fluctuations in metric scores for all methods. This indicates that most methods, including ours, exhibit strong robustness to inference step changes. 

\subsection{Comparison on Strength}

In this section, we evaluate the inpainting performance of various methods by varying the strength parameter from 1.0 to 0.8. As described in Sec. \ref{Subsec: Inpainting background}, when the strength is set to 1.0, the latent variable $z_T$ is initialized as pure random noise. In this setting, information about the original image is accessible only through the clean masked latent $z_0^\mathbf{M}$, which lacks contextual cues about the masked region. However, when the strength is set below 1.0, $z_T$ is generated by applying $T$ steps of Gaussian noise to $z_0$ and, as such, is no longer entirely random. This process introduces a partial image-level context that may inform the model about the masked area.

This effect is particularly pronounced in the EditBench evaluation dataset, where the masked regions in the original images align closely with the prompt content. Consequently, reducing the strength increases the likelihood of content generation within the masked regions, as the image-level and prompt-level contexts become mutually reinforcing.

\begin{figure*}[th]
  \centering
  \includegraphics[width=\linewidth]{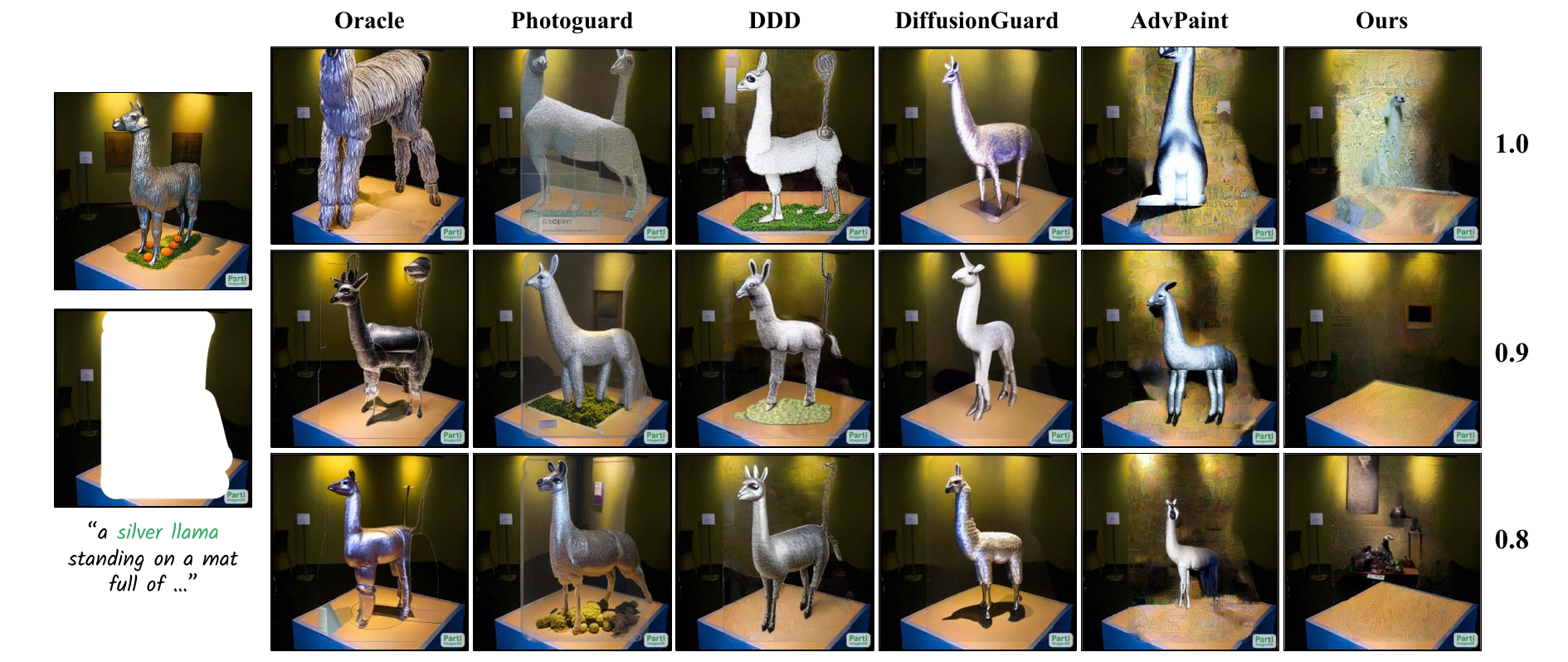}
  \caption{Qualitative evaluation of various methods on the EditBench dataset across different strengths. Unlike baseline methods, which increasingly converge to a specific llama appearance as strength decreases, our method effectively suppresses content generation regardless of the strength level.}
  \Description{A woman and a girl in white dresses sit in an open car.}
  \label{AFig: strength}
\end{figure*}

Fig. \ref{AFig: strength} supports this analysis by qualitatively comparing the inpainting results across varying strength values. As noted previously, when the strength is set to 1.0, the baseline methods produce diverse renditions of a llama, reflecting their exclusive dependence on the prompt. However, as the strength decreases, the generated outputs increasingly resemble the llamas in the original images, indicating a greater influence from the image-level context. Despite the growing challenge of suppressing content in such scenarios, our method consistently achieves successful generation suppression.

Tab. \ref{ATab: strength} provides a quantitative comparison across different strength settings. In line with prior experiments, our method exhibits significant margins in CLIP Score across all strength levels and consistently achieves top performance across most evaluation metrics.

\subsection{Robustness}

In this section, we assess the robustness of various methods against attacks intended to remove adversarial noise, which may be used to circumvent generation suppression for malicious manipulation. For each method, we apply five types of attacks, including AdverseCleaner~\cite{Zhang2025AdverseCleaner} and JPEG~\cite{wallace1991jpeg} compression with quality levels ranging from 90 to 60, and compare the resulting inpainting outputs.

\begin{figure*}[th]
  \centering
  \includegraphics[width=\linewidth]{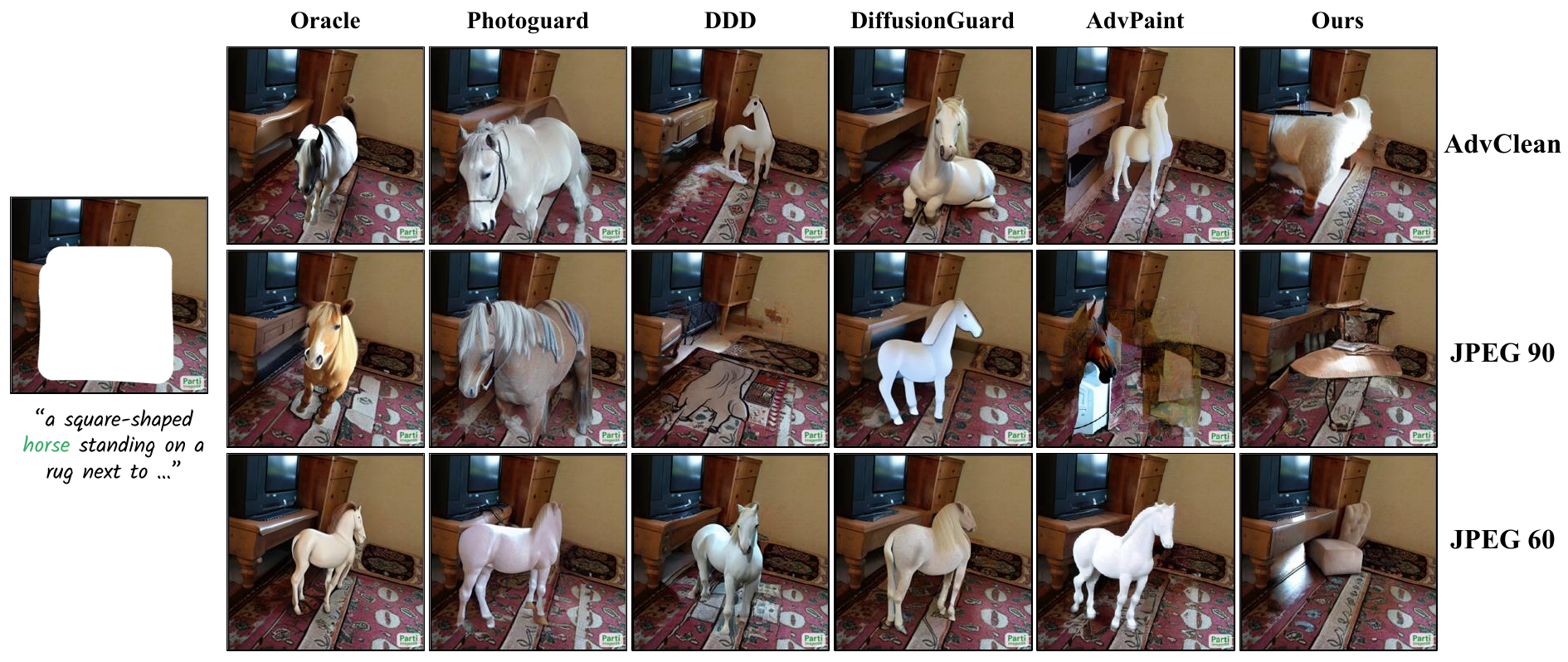}
  \caption{Qualititive evaluation of various methods on the EditBench dataset across various adversarial noise removal attacks. While baseline methods fail to suppress content under AdverseCleaner and JPEG compression, our method demonstrates enhanced robustness, maintaining effective generation suppression even under severe conditions.}
  \Description{A woman and a girl in white dresses sit in an open car.}
  \label{AFig: robustness}
\end{figure*}

Fig. \ref{AFig: robustness} illustrates a qualitative comparison of inpainting results under different adversarial noise removal attacks. AdverseCleaner proves to be effective, causing all baseline methods to fail in suppressing content generation. In contrast to previous experiment result, even our method produces a furred object, although it is far from being recognizable as a horse. JPEG compression attacks significantly impair the baseline methods, while our method demonstrates substantially greater robustness.

Tab. \ref{ATab: robustness} supports these findings through quantitative analysis. Across all types of attack, our method consistently achieves the highest performance in nearly all reference-free metrics. Notably, while baseline methods already approach Oracle level performance under JPEG compression with quality 80, our method continues to maintain a noticeable gap from the Oracle even at quality 60 in reference-free metrics. It highlights its sustained suppression effectiveness under stronger attack scenarios.

\begin{table*}[th]
    \centering
    \caption{Quantitative evaluation of various methods on the EditBench dataset across different inference steps. Evaluation metrics and all parameters except for the inference step follow the configurations specified in the main paper. The best results are highlighted in bold, and the second-best results are underlined. Our method consistently achieves the best performance in nearly all settings.}
    \label{ATab: inference step}
    \begin{tabular}{c | c c c c c c c c c c}
    \toprule
    inference
        & \multirow{2}{*}{Methods}
        & \multicolumn{2}{c}{CLIP Score $\downarrow$}
        & \multicolumn{2}{c}{Aesthetic $\downarrow$}
        & \multicolumn{2}{c}{PickScore $\downarrow$}
        & \multirow{2}{*}{LPIPS $\uparrow$}
        & \multirow{2}{*}{SSIM $\downarrow$}
        & \multirow{2}{*}{PSNR $\downarrow$} \\
    \cmidrule(lr){3-4}
    \cmidrule(lr){5-6}
    \cmidrule(lr){7-8}
    step & & \textit{all} & \textit{mask}
    & \textit{all} & \textit{mask}
    & \textit{all} & \textit{mask}
    &  &  &  \\
    \midrule
    \multirow{5}{*}{100} & Oracle
        & 32.1122 & 26.6906 & 4.8256 & 4.4101 & 20.7652 & 19.6904
        & --      & --      & --      \\
    & PhotoGuard~\cite{salman2023raising}
        & 32.0911 & 26.8208 & 4.5739 & 4.3302 & 20.4813 & 19.6682 & 0.2420 & 0.6937 & 14.6586 \\
    & DDD~\cite{son2024disrupting}
        & 31.1849 & 25.5692 & 4.5707 & 4.2979 & 20.2534 & 19.3904 & 0.2428 & 0.6947 & \textbf{14.1971} \\
    & DiffusionGuard~\cite{choi2024diffusionguard}
        & 30.6389 & 24.9987 & 4.6266 & 4.1885 & 20.3724 & 19.2744 & 0.2540 & 0.6981 & 14.3685 \\
    & AdvPaint~\cite{jeon2025advpaint}
        & \underline{28.8135} & \underline{22.8955} & \textbf{4.3444} & \underline{4.1916} & \underline{19.6349} & \underline{18.8107} & \textbf{0.2660} & \textbf{0.6830} & \underline{14.2166} \\
    & \cellcolor{backcolour}\SystemName~(\textit{\textbf{ours}})
        & \cellcolor{backcolour}\textbf{27.3124} & \cellcolor{backcolour}\textbf{21.8493} & \cellcolor{backcolour}\underline{4.4703} & \cellcolor{backcolour}\textbf{4.0965} & \cellcolor{backcolour}\textbf{19.5790} & \cellcolor{backcolour}\textbf{18.6325} & \cellcolor{backcolour}\underline{0.2557} & \cellcolor{backcolour}\underline{0.6884} & \cellcolor{backcolour}14.8475 \\
    \midrule
    \multirow{5}{*}{75} & Oracle
        & 32.3319 & 26.8320 & 4.8136 & 4.3954 & 20.7430 & 19.6942
        & --      & --      & --      \\
    & PhotoGuard~\cite{salman2023raising}
        & 31.8356 & 26.5574 & 4.5609 & 4.2859 & 20.4497 & 19.6538 & 0.2421 & 0.6911 & 14.5638 \\
    & DDD~\cite{son2024disrupting}
        & 31.0712 & 25.3132 & 4.5662 & 4.3089 & 20.2787 & 19.4051 & 0.2441 & 0.6921 & \underline{14.0539} \\
    & DiffusionGuard~\cite{choi2024diffusionguard}
        & 30.5320 & 25.0023 & 4.6141 & 4.1882 & 20.3229 & 19.2156 & \underline{0.2558} & 0.6946 & 14.2036 \\
    & AdvPaint~\cite{jeon2025advpaint}
        & \underline{28.7885} & \underline{22.8615} & \textbf{4.3230} & \underline{4.1655} & \underline{19.6306} & \underline{18.8219} & \textbf{0.2662} & \textbf{0.6819} & \textbf{14.0504} \\
    & \cellcolor{backcolour}\SystemName~(\textit{\textbf{ours}})
        & \cellcolor{backcolour}\textbf{27.5925} & \cellcolor{backcolour}\textbf{22.1424} & \cellcolor{backcolour}\underline{4.4665} & \cellcolor{backcolour}\textbf{4.1099} & \cellcolor{backcolour}\textbf{19.5925} & \cellcolor{backcolour}\textbf{18.6744} & \cellcolor{backcolour}0.2555 & \cellcolor{backcolour}\underline{0.6867} & \cellcolor{backcolour}14.7293 \\
        \midrule
    \multirow{5}{*}{50} & Oracle
        & 32.0412 & 26.6546
        & 4.7981  & 4.3493
        & 20.7639 & 19.6893
        & --      & --      & --      \\
    & PhotoGuard~\cite{salman2023raising}
        & 31.7589 & 26.3759
        & 4.5361  & 4.2780
        & 20.4062 & 19.6097
        & 0.2443  & 0.6947  & 14.8245 \\
    & DDD~\cite{son2024disrupting}
        & 30.9199 & 25.2935
        & 4.5625  & 4.5685
        & 20.2364 & 19.3621
        & 0.2449  & 0.6969  & \textbf{14.2616} \\
    & DiffusionGuard~\cite{choi2024diffusionguard}
        & 30.6709 & 25.1593
        & 4.6233  & 4.2097
        & 20.3921 & 19.3010
        & 0.2547  & 0.7016  & 14.5926 \\
    & AdvPaint~\cite{jeon2025advpaint}
        & \underline{28.4993} & \underline{22.8013} & \textbf{4.3205} & \underline{4.1715} & \underline{19.6327} & \underline{18.8247} & \textbf{0.2683} & \textbf{0.6839} & \underline{14.4043} \\
    & \cellcolor{backcolour}\SystemName~(\textit{\textbf{ours}})
        & \cellcolor{backcolour}\textbf{27.3171} & \cellcolor{backcolour}\textbf{21.9557}
        & \cellcolor{backcolour}\underline{4.4569}  & \cellcolor{backcolour}\textbf{4.0916}
        & \cellcolor{backcolour}\textbf{19.5498} & \cellcolor{backcolour}\textbf{18.6491}
        & \cellcolor{backcolour}\underline{0.2570}  & \cellcolor{backcolour}\underline{0.6894}  & \cellcolor{backcolour}14.9311 \\
    \bottomrule
    \end{tabular}
\end{table*}

\begin{table*}[th]
    \centering
    \caption{Quantitative evaluation of various methods on the EditBench dataset across different strength. Evaluation metrics and all parameters except for the strength follow the configurations specified in the main paper. The best results are highlighted in bold, and the second-best results are underlined. Our method consistently achieves the best performance in nearly all settings.}
    \label{ATab: strength}
    \begin{tabular}{c | c c c c c c c c c c}
    \toprule
    \multirow{2}{*}{strength}
        & \multirow{2}{*}{Methods}
        & \multicolumn{2}{c}{CLIP Score $\downarrow$}
        & \multicolumn{2}{c}{Aesthetic $\downarrow$}
        & \multicolumn{2}{c}{PickScore $\downarrow$}
        & \multirow{2}{*}{LPIPS $\uparrow$}
        & \multirow{2}{*}{SSIM $\downarrow$}
        & \multirow{2}{*}{PSNR $\downarrow$} \\
    \cmidrule(lr){3-4}
    \cmidrule(lr){5-6}
    \cmidrule(lr){7-8}
    & & \textit{all} & \textit{mask}
    & \textit{all} & \textit{mask}
    & \textit{all} & \textit{mask}
    &  &  &  \\
    \midrule
    \multirow{5}{*}{1.0} & Oracle
        & 32.0412 & 26.6546
        & 4.7981  & 4.3493
        & 20.7639 & 19.6893
        & --      & --      & --      \\
    & PhotoGuard~\cite{salman2023raising}
        & 31.7589 & 26.3759
        & 4.5361  & 4.2780
        & 20.4062 & 19.6097
        & 0.2443  & 0.6947  & 14.8245 \\
    & DDD~\cite{son2024disrupting}
        & 30.9199 & 25.2935
        & 4.5625  & 4.5685
        & 20.2364 & 19.3621
        & 0.2449  & 0.6969  & \textbf{14.2616} \\
    & DiffusionGuard~\cite{choi2024diffusionguard}
        & 30.6709 & 25.1593
        & 4.6233  & 4.2097
        & 20.3921 & 19.3010
        & 0.2547  & 0.7016  & 14.5926 \\
    & AdvPaint~\cite{jeon2025advpaint}
        & \underline{28.4993} & \underline{22.8013} & \textbf{4.3205} & \underline{4.1715} & \underline{19.6327} & \underline{18.8247} & \textbf{0.2683} & \textbf{0.6839} & \underline{14.4043} \\
    & \cellcolor{backcolour}\SystemName~(\textit{\textbf{ours}})
        & \cellcolor{backcolour}\textbf{27.3171} & \cellcolor{backcolour}\textbf{21.9557}
        & \cellcolor{backcolour}\underline{4.4569}  & \cellcolor{backcolour}\textbf{4.0916}
        & \cellcolor{backcolour}\textbf{19.5498} & \cellcolor{backcolour}\textbf{18.6491}
        & \cellcolor{backcolour}\underline{0.2570}  & \cellcolor{backcolour}\underline{0.6894}  & \cellcolor{backcolour}14.9311 \\
    \midrule
    \multirow{5}{*}{0.9} & Oracle
        & 32.3348 & 26.8842 & 4.7935 & 4.3486 & 20.8546 & 19.7076
        & --      & --      & --      \\
    & PhotoGuard~\cite{salman2023raising}
        & 31.9582 & 26.7666 & 4.5773 & 4.2762 & 20.4295 & 19.6322 & 0.2320 & {0.7271} & 16.2356 \\
    & DDD~\cite{son2024disrupting}
        & 31.3280 & 25.5532 & 4.5458 & 4.2237 & 20.3420 & 19.4522 & 0.2277 & 0.7350 & \underline{15.9755} \\
    & DiffusionGuard~\cite{choi2024diffusionguard}
        & 30.7107 & 25.3450 & 4.6447 & 4.1705 & 20.3847 & 19.3301 & 0.2369 & 0.7408 & 16.1842 \\
    & AdvPaint~\cite{jeon2025advpaint}
        & \underline{28.9641} & \underline{23.2302} & \textbf{4.3391} & \underline{4.1425} & \underline{19.7477} & \underline{18.9107} & \textbf{0.2598} & \textbf{0.7091} & \textbf{15.8875} \\
    & \cellcolor{backcolour}\SystemName~(\textit{\textbf{ours}})
        & \cellcolor{backcolour}\textbf{27.9966} & \cellcolor{backcolour}\textbf{22.6204} & \cellcolor{backcolour}\underline{4.4773} & \cellcolor{backcolour}\textbf{4.1093} & \cellcolor{backcolour}\textbf{19.7417} & \cellcolor{backcolour}\textbf{18.7737} & \cellcolor{backcolour}\underline{0.2432} & \cellcolor{backcolour}\underline{0.7266} & \cellcolor{backcolour}16.6115 \\
        \midrule
    \multirow{5}{*}{0.8} & Oracle
        & 32.6008 & 27.1339 & 4.7674 & 4.3120 & 20.8970 & 19.7893
        & --      & --      & --      \\
    & PhotoGuard~\cite{salman2023raising}
        & 32.3497 & 26.8865 & 4.5460 & 4.2572 & 20.5697 & 19.7289 & 0.2231 & 0.7413 & 17.1554 \\
    & DDD~\cite{son2024disrupting}
        & 31.6128 & 26.1634 & 4.5741 & 4.2455 & 20.4488 & 19.5193 & 0.2178 & 0.7502 & \underline{17.0701} \\
    & DiffusionGuard~\cite{choi2024diffusionguard}
        & 31.2720 & 25.9908 & 4.6793 & 4.2147 & 20.5487 & 19.4751 & 0.2251 & 0.7543 & 17.3044 \\
    & AdvPaint~\cite{jeon2025advpaint}
        & \underline{29.9672} & \underline{24.0975} & \textbf{4.3380} & \underline{4.1709} & \underline{19.9273} & \underline{19.0865} & \textbf{0.2500} & \textbf{0.7229} & \textbf{16.8790} \\
    & \cellcolor{backcolour}\SystemName~(\textit{\textbf{ours}})
        & \cellcolor{backcolour}\textbf{28.8508} & \cellcolor{backcolour}\textbf{23.3454} & \cellcolor{backcolour}\underline{4.5009} & \cellcolor{backcolour}\textbf{4.1304} & \cellcolor{backcolour}\textbf{19.8919} & \cellcolor{backcolour}\textbf{18.9394} & \cellcolor{backcolour}\underline{0.2338} & \cellcolor{backcolour}\underline{0.7394} & \cellcolor{backcolour}17.6467 \\
        \bottomrule
    \end{tabular}
\end{table*}

\begin{table*}[th]
    \centering
    \caption{Quantitative evaluation of various methods on the EditBench dataset across various attacks. Evaluation metrics and all parameters except for the strength follow the configurations specified in the main paper. The best results are highlighted in bold, and the second-best results are underlined. Our method consistently achieves the best performance in nearly all settings.}
    \label{ATab: robustness}
    \begin{tabular}{c | c c c c c c c c c c}
    \toprule
    \multirow{2}{*}{Attack}
        & \multirow{2}{*}{Methods}
        & \multicolumn{2}{c}{CLIP Score $\downarrow$}
        & \multicolumn{2}{c}{Aesthetic $\downarrow$}
        & \multicolumn{2}{c}{PickScore $\downarrow$}
        & \multirow{2}{*}{LPIPS $\uparrow$}
        & \multirow{2}{*}{SSIM $\downarrow$}
        & \multirow{2}{*}{PSNR $\downarrow$} \\
    \cmidrule(lr){3-4}
    \cmidrule(lr){5-6}
    \cmidrule(lr){7-8}
    & & \textit{all} & \textit{mask}
    & \textit{all} & \textit{mask}
    & \textit{all} & \textit{mask}
    &  &  &  \\
    \midrule
    & Oracle
        & 31.9432 & 26.6936 & 4.8592 & 4.4171 & 20.8687 & 19.7181
        & --      & --      & --      \\
    & PhotoGuard~\cite{salman2023raising}
        & 32.2047 & 26.7201 & \textbf{4.7387} & 4.3751 & 20.8288 & 19.7168 & 0.2215 & \underline{0.7334} & 14.7483 \\
    Adverse & DDD~\cite{son2024disrupting}
        & 32.0131 & 26.5687 & 4.7980 & 4.3381 & 20.7745 & 19.6234 & \underline{0.2220} & \textbf{0.7323} & \textbf{14.5317} \\
    Cleaner & DiffusionGuard~\cite{choi2024diffusionguard}
        & \underline{31.5614} & \underline{26.1341} & 4.7687 & \underline{4.3198} & \underline{20.6886} & \underline{19.5486} & 0.2177 & 0.7422 & 14.8911 \\
    & AdvPaint~\cite{jeon2025advpaint}
        & 31.8201 & 26.3814 & 4.7748 & 4.3680 & 20.7011 & 19.6109 & \textbf{0.2251} & 0.7346 & \underline{14.5707} \\
    & \cellcolor{backcolour}\SystemName~(\textit{\textbf{ours}})
        & \cellcolor{backcolour}\textbf{30.9129} & \cellcolor{backcolour}\textbf{25.6647} & \cellcolor{backcolour}\underline{4.7570} & \cellcolor{backcolour}\textbf{4.2949} & \cellcolor{backcolour}\textbf{20.5889} & \cellcolor{backcolour}\textbf{19.4570} & \cellcolor{backcolour}0.2182 & \cellcolor{backcolour}0.7376 & \cellcolor{backcolour}14.6657 \\
    \midrule
    \multirow{5}{*}{JPEG 90} & Oracle
        & 31.9618 & 26.3973 & 4.8511 & 4.3600 & 20.8583 & 19.6833
        & --      & --      & --      \\
    & PhotoGuard~\cite{salman2023raising}
        & 32.1022 & 26.8707 & 4.6200 & \underline{4.1782} & 20.6940 & 19.6432 & 0.2353 & \underline{0.7155} & 14.8834 \\
    & DDD~\cite{son2024disrupting}
        & 31.3873 & 25.8143 & 4.6230 & 4.2035 & 20.4828 & 19.4597 & \underline{0.2398} & 0.7161 & \textbf{14.5640} \\
    & DiffusionGuard~\cite{choi2024diffusionguard}
        & 31.0219 & 25.6764 & 4.7495 & 4.2328 & 20.5557 & 19.4356 & 0.2266 & 0.7313 & 14.8050 \\
    & AdvPaint~\cite{jeon2025advpaint}
        & \underline{30.5213} & \underline{24.7324} & \textbf{4.4515} & 4.2005 & \underline{20.1618} & \underline{19.2136} & \textbf{0.2517} & \textbf{0.7102} & \underline{14.5985} \\
    & \cellcolor{backcolour}\SystemName~(\textit{\textbf{ours}})
        & \cellcolor{backcolour}\textbf{28.9882} & \cellcolor{backcolour}\textbf{23.6347} & \cellcolor{backcolour}\underline{4.5923} & \cellcolor{backcolour}\textbf{4.1443} & \cellcolor{backcolour}\textbf{20.0484} & \cellcolor{backcolour}\textbf{19.0151} & \cellcolor{backcolour}0.2362 & \cellcolor{backcolour}0.7197 & \cellcolor{backcolour}15.0224 \\
        \midrule
    \multirow{5}{*}{JPEG 80} & Oracle
        & 31.6928 & 26.4773 & 4.8357 & 4.3430 & 20.8388 & 19.6551
        & --      & --      & --      \\
    & PhotoGuard~\cite{salman2023raising}
        & 31.9140 & 26.7487 & 4.7407 & 4.1755 & 20.8276 & 19.6801 & 0.2223 & 0.7268 & 14.9739 \\
    & DDD~\cite{son2024disrupting}
        & 31.8344 & 26.5383 & 4.7406 & 4.2035 & 20.7493 & 19.6237 & \underline{0.2282} & \textbf{0.7248} & \underline{14.8041} \\
    & DiffusionGuard~\cite{choi2024diffusionguard}
        & \underline{31.5472} & \underline{26.3611} & 4.7612 & \textbf{4.1497} & 20.7409 & 19.6310 & 0.2182 & 0.7370 & 15.0993 \\
    & AdvPaint~\cite{jeon2025advpaint}
        & 31.9592 & 26.4586 & \underline{4.6985} & 4.1818 & \underline{20.7352} & \underline{19.5840} & \textbf{0.2327} & \underline{0.7253} & \textbf{14.7385} \\
    & \cellcolor{backcolour}\SystemName~(\textit{\textbf{ours}})
        & \cellcolor{backcolour}\textbf{30.6442} & \cellcolor{backcolour}\textbf{24.9441} & \cellcolor{backcolour}\textbf{4.6865} & \cellcolor{backcolour}\underline{4.1602} & \cellcolor{backcolour}\textbf{20.4508} & \cellcolor{backcolour}\textbf{19.2816} & \cellcolor{backcolour}0.2254 & \cellcolor{backcolour}0.7298 & \cellcolor{backcolour}15.0526 \\
        \midrule
    \multirow{5}{*}{JPEG 70} & Oracle
        & 31.9409 & 26.6245 & 4.8531 & 4.3414 & 20.7843 & 19.6441
        & --      & --      & --      \\
    & PhotoGuard~\cite{salman2023raising}
        & 32.0063 & 26.7420 & 4.7762 & 4.1853 & 20.8137 & 19.6250 & 0.2176 & 0.7312 & 14.8894 \\
    & DDD~\cite{son2024disrupting}
        & 32.3094 & 26.8395 & 4.8043 & 4.2241 & 20.8318 & 19.6709 & \underline{0.2235} & \textbf{0.7256} & \underline{14.6335} \\
    & DiffusionGuard~\cite{choi2024diffusionguard}
        & \underline{31.5901} & \underline{26.1541} & 4.8093 & 4.1747 & \underline{20.7176} & \underline{19.5683} & 0.2157 & 0.7346 & 14.9020 \\
    & AdvPaint~\cite{jeon2025advpaint}
        & 31.8491 & 26.5589 & \underline{4.7541} & \underline{4.1602} & 20.7639 & 19.5964 & \textbf{0.2257} & \underline{0.7264} & \textbf{14.5715} \\
    & \cellcolor{backcolour}\SystemName~(\textit{\textbf{ours}})
        & \cellcolor{backcolour}\textbf{31.1178} & \cellcolor{backcolour}\textbf{25.5631} & \cellcolor{backcolour}\textbf{4.7408} & \cellcolor{backcolour}\textbf{4.1598} & \cellcolor{backcolour}\textbf{20.6318} & \cellcolor{backcolour}\textbf{19.4498} & \cellcolor{backcolour}0.2209 & \cellcolor{backcolour}0.7303 & \cellcolor{backcolour}14.6593 \\
        \midrule
    \multirow{5}{*}{JPEG 60} & Oracle
        & \underline{31.7861} & \underline{26.3823} & 4.8498 & 4.3182 & 20.8224 & 19.6529
        & --      & --      & --      \\
    & PhotoGuard~\cite{salman2023raising}
        & 31.8820 & 26.5382 & 4.7867 & 4.1644 & 20.8128 & 19.6729 & 0.2161 & 0.7317 & 14.8920 \\
    & DDD~\cite{son2024disrupting}
        & 32.3947 & 27.0525 & 4.8184 & 4.2370 & 20.9346 & 19.7697 & \textbf{0.2197} & \textbf{0.7293} & \textbf{14.6581} \\
    & DiffusionGuard~\cite{choi2024diffusionguard}
        & 31.8285 & 26.6167 & 4.7979 & 4.1897 & \underline{20.7515} & \underline{19.5746} & 0.2147 & 0.7359 & 14.9816 \\
    & AdvPaint~\cite{jeon2025advpaint}
        & 31.9259 & 26.5758 & \textbf{4.7554} & \textbf{4.0957} & 20.7973 & 19.5861 & \underline{0.2178} & \underline{0.7321} & \underline{14.8201} \\
    & \cellcolor{backcolour}\SystemName~(\textit{\textbf{ours}})
        & \cellcolor{backcolour}\textbf{31.4824} & \cellcolor{backcolour}\textbf{25.8161} & \cellcolor{backcolour}\underline{4.7598} & \cellcolor{backcolour}\underline{4.1191} & \cellcolor{backcolour}\textbf{20.6363} & \cellcolor{backcolour}\textbf{19.4305} & \cellcolor{backcolour}0.2169 & \cellcolor{backcolour}0.7326 & \cellcolor{backcolour}14.8606 \\
        \bottomrule
    \end{tabular}
\end{table*}

\end{document}